\documentclass[10pt]{article}

\usepackage{times}
\usepackage[margin=1in]{geometry}
\usepackage[round,compress]{natbib}
\usepackage{parskip}

\usepackage{parskip}
\usepackage{subcaption}
\usepackage[font=small,labelfont=bf]{caption}

\usepackage{amsmath}
\usepackage{amsfonts,amscd,amssymb,bm,bbm,mathrsfs}
\usepackage{nicefrac}       
\usepackage{amsthm}
\usepackage{mathtools}

\newcommand\labelAndRemember[2]
  {\expandafter\gdef\csname labeled:#1\endcsname{#2}%
   \label{#1}#2}
\newcommand\recallLabel[1]
   {\csname labeled:#1\endcsname\tag{\ref{#1}}}
\newcommand\labelr[2]
  {\expandafter\gdef\csname labeled:#1\endcsname{#2}%
   \label{#1}#2}
\newcommand\recall[1]
   {\csname labeled:#1\endcsname}

\usepackage[algo2e,linesnumbered,vlined,ruled]{algorithm2e}
\usepackage{algorithm}
\usepackage[noend]{algorithmic}

\usepackage{url}

\usepackage[titletoc,title]{appendix}
\usepackage{enumerate}
\usepackage{enumitem}
\usepackage{comment}

\usepackage{booktabs}       
\usepackage{multirow}
\usepackage{multicol}
\usepackage{makecell}
\usepackage{array}
\usepackage{arydshln}
\newcolumntype{H}{>{\setbox0=\hbox\bgroup}c<{\egroup}@{}}
\newcolumntype{Z}{>{\setbox0=\hbox\bgroup}c<{\egroup}@{\hspace*{-\tabcolsep}}}

\usepackage[normalem]{ulem}
\usepackage{exscale}
\usepackage{tikz}
\usepackage{float}
\usepackage{graphicx,graphics}
\graphicspath{ {./fig/} }
\allowdisplaybreaks

\newtheorem{theorem}{Theorem}
\newtheorem*{theorem*}{Theorem}
\newtheorem{lemma}[theorem]{Lemma}

\newtheorem*{remark*}{Remark}
\newtheorem*{lemma*}{Lemma}

\newtheorem{proposition}[theorem]{Proposition}

\newenvironment{proof-sketch}{\noindent{\bf Proof Sketch}
  \hspace*{1em}}{\qed\bigskip\\}
\newenvironment{proof-idea}{\noindent{\bf Proof Idea}
  \hspace*{1em}}{\qed\bigskip\\}
\newenvironment{proof-of}[1][{}]{\noindent{\bf Proof of \cref{#1}}
  \hspace*{1em}}{\qed\bigskip\\}
\newenvironment{proof-of-lemma}[1][{}]{\noindent{\bf Proof of Lemma {#1}}
  \hspace*{1em}}{\qed\bigskip\\}
\newenvironment{proof-of-proposition}[1][{}]{\noindent{\bf
    Proof of Proposition {#1}}
  \hspace*{1em}}{\qed\bigskip\\}
\newenvironment{proof-of-theorem}[1][{}]{\noindent{\bf Proof of Theorem {#1}}
  \hspace*{1em}}{\qed\bigskip\\}
\newenvironment{inner-proof}{\noindent{\bf Proof}\hspace{1em}}{
  $\bigtriangledown$\medskip\\}
\newenvironment{proof-attempt}{\noindent{\bf Proof Attempt}
  \hspace*{1em}}{\qed\bigskip\\}

\renewcommand{\epsilon}{\varepsilon}

\usepackage{pifont}

\newcommand{\eps}{\varepsilon}

\newcounter{cnt}
\foreach \num in {1,2,...,26}{%
  \stepcounter{cnt}%
  \expandafter\xdef \csname c\Alph{cnt}\endcsname {\noexpand\mathcal{\Alph{cnt}}}%
  \expandafter\xdef \csname b\Alph{cnt}\endcsname {\noexpand\mathbb{\Alph{cnt}}}%
}

\newcommand{\diag}{\operatorname{diag}}

\DeclarePairedDelimiterX{\ddiv}[2]{(}{)}{%
  #1\;\delimsize\|\;#2%
}

\newcommand{\norm}[1]{\left\|{#1}\right\|} 
\newcommand{\ltwo}[1]{\norm{#1}_2} 

\newcommand{\<}{\left\langle}
\renewcommand{\>}{\right\rangle}

\newenvironment{talign*}
 {\csname align*\endcsname}
 {\endalign}
 
\newcommand{\zeroth}{{(0)}}

\mathchardef\mhyphen="2D

\newcommand{\sigmoid}{{\rm sigmoid}}

\newcommand{\pred}{\mathrm{pred}}

\newcommand{\brac}[1]{{\left[ #1 \right]}}

\newcommand{\sets}[1]{{\{ #1 \}}}

\newcommand{\E}{\mathbb{E}}

\renewcommand{\P}{\mathbb{P}}

\newcommand{\R}{\mathbb{R}}

\usepackage{amsmath,amsfonts,bm}

\def\eqref#1{equation~\ref{#1}}

\def\1{\bm{1}}

\def\eps{{\epsilon}}

\DeclareMathAlphabet{\mathsfit}{\encodingdefault}{\sfdefault}{m}{sl}
\SetMathAlphabet{\mathsfit}{bold}{\encodingdefault}{\sfdefault}{bx}{n}

\def\gN{{\mathcal{N}}}

\newcommand{\softmax}{\mathrm{softmax}}

\newcommand{\Prob}{{\mathbb{P}}}
\newcommand{\Dim}{{d}}
\newcommand{\ones}{{\mathbf{1}}}
\newcommand{\zeros}{{\mathbf{0}}}

\newcommand{\NPO}{{\mathrm{NPO}}}
\newcommand{\DPO}{{\mathrm{DPO}}}
\newcommand{\GA}{{\mathrm{GA}}}

\newcommand{\IDK}{{\mathrm{IDK}}}

\newcommand{\FG}{{\mathrm{FG}}}
\newcommand{\RT}{{\mathrm{RT}}}

\newcommand{\xforget}{x}

\newcommand{\yforget}{y}

\newcommand{\nforget}{{n_{\mathrm f}}}

\newcommand{\exforget}{x^{\mathrm f}}
\newcommand{\exretain}{x^{\mathrm r}}
\newcommand{\eyforget}{y^{\mathrm f}}
\newcommand{\eyretain}{y^{\mathrm r}}

\newcommand{\clga}{c_{\rm GA}}
\newcommand{\clfg}{c_{\rm FG}}
\newcommand{\clrt}{c_{\rm RT}}
\newcommand{\ckfg}{c_{\rm FGKL}}
\newcommand{\ckrt}{c_{\rm RTKL}}

\newcommand{\policy}{{\pi}}
\newcommand{\Par}{{\theta}}

\newcommand{\Loss}{\mathcal{L}}
\newcommand{\dset}{{\mathcal{D}}}
\newcommand{\kldivergence}{{\mathsf {D}}}
\newcommand{\Lossdiv}{{\mathcal{K}}}

\newcommand{\dsetf}{{\dset_{\rm FG}}}
\newcommand{\dsetr}{{\dset_{\rm RT}}}
\newcommand{\itemp}{{\beta}}
\newcommand{\refm}{{\mathrm{ref}}}
\newcommand{\const}{c}  
\newcommand{\pconst}{c} 

\newcommand{\retrain}{\rm retr}
\newcommand{\ReLU}{\rm ReLU}
\newcommand{\paired}{\rm paired}
\newcommand{\good}{\rm w}
\newcommand{\bad}{\rm l}

\newcommand{\x}{{x}}
\newcommand{\y}{{y}}
\newcommand{\sigmoidshort}{{\sigma}}
\newcommand{\logratio}{{\mathsf{R}}}
\newcommand{\NPOweight}{\mathsf{W}}
\newcommand{\init}{\mathrm{init}}
\newcommand{\tottime}{T}
\newcommand{\tth}{{(t)}}

\newcommand{\Kth}{{(K)}}
\newcommand{\Kponeth}{{(K+1)}}

\newcommand{\first}{{(1)}}

\newcommand{\tponeth}{{(t+1)}}

\newcommand{\corre}{{\gamma}}
\newcommand{\Corre}{{\Gamma}}

\newcommand{\direc}{{\alpha}}
\newcommand{\lrate}{{\eta_0}}  
\newcommand{\lrateinit}{{\eta}} 

\newcommand{\direcb}{{b}}

\newcommand{\polyshort}{{C}}

\newcommand{\boundx}{{B_x}}
\newcommand{\lboundx}{{b_x}}

\newcommand{\boundinit}{{B_\Par}}

\newcommand{\diff}{{\Delta}} 
\newcommand{\epsth}{{(\eps)}}
\newcommand{\Xforget}{{X}}

\usepackage{hyperref}
\usepackage{cleveref}

\title{Negative Preference Optimization: From Catastrophic  Collapse \\ to Effective Unlearning}

\date{\today}
\author{
  Ruiqi Zhang\thanks{Equal contributions; the more junior author is listed earlier.} \ \thanks{UC Berkeley. Email: \texttt{rqzhang@berkeley.edu}}\hspace{.35em}
  \and
  Licong Lin\footnotemark[1] \ \thanks{UC Berkeley. Email: \texttt{liconglin@berkeley.edu}}\hspace{.35em}
  \and
  Yu Bai\thanks{Salesforce AI Research. Email: \texttt{yu.bai@salesforce.com}}\hspace{.35em}
  \and
  Song Mei\thanks{UC Berkeley. Email: \texttt{songmei@berkeley.edu}}\hspace{.35em}
}

\def\shownotes{0}  
\ifnum\shownotes=1
\newcommand{\authnote}[2]{{\scriptsize $\ll$\textsf{#1 notes: #2}$\gg$}}
\else
\newcommand{\authnote}[2]{}
\fi

\makeatletter
\def\blfootnote{\gdef\@thefnmark{}\@footnotetext}
\makeatother

\begin{document}

\maketitle

\date{}

\begin{abstract}
Large Language Models (LLMs) often memorize sensitive, private, or copyrighted data during pre-training. LLM unlearning aims to eliminate the influence of undesirable data from the pre-trained model while preserving the model's utilities on other tasks. Several practical methods have recently been proposed for LLM unlearning, mostly based on gradient ascent (GA) on the loss of undesirable data. However, on certain unlearning tasks, these methods either fail to effectively unlearn the target data or suffer from catastrophic collapse---a drastic degradation of the model's utilities. 

In this paper, we propose \emph{Negative Preference Optimization} (NPO), a simple alignment-inspired method that could efficiently and effectively unlearn a target dataset. We theoretically show that the progression toward catastrophic collapse by minimizing the NPO loss is exponentially slower than GA. Through experiments
on synthetic data and the benchmark TOFU dataset, we demonstrate that NPO-based methods achieve a better balance between unlearning the undesirable data and maintaining the model's utilities. 
We also observe that NPO-based methods generate more sensible outputs than GA-based methods, whose outputs are often gibberish.
Remarkably, on TOFU, NPO-based methods are the first to achieve reasonable unlearning results in forgetting 50\% (or more) of the training data, whereas existing methods already struggle with forgetting 10\% of training data. \blfootnote{Code is available at:  \href{https://github.com/licong-lin/negative-preference-optimization}{https://github.com/licong-lin/negative-preference-optimization}.}

\end{abstract}
\section{Introduction}

Large language models (LLMs), pretrained on massive corpora of internet data, possess the capability to memorize portions of their training data \citep{carlini2021extracting, carlini2022quantifying}. However, this capability raises significant concerns, as the training data may contain sensitive or private information, potentially leading to societal challenges. For instance, language models could breach individual privacy by outputting personal information such as social security numbers from the memorized data \citep{carlini2021extracting, huang2022large}. They might also violate copyright by generating text from memorized books, such as the Harry Potter novels \citep{eldan2023s}. Furthermore, LLM assistants for biology could inadvertently aid in the development of biological weapons by troubleshooting bottlenecks, increasing the risk of such attempts \citep{sandbrink2023artificial, li2024wmdp}. In response to these concerns, regulations like the EU's General Data Protection Regulation (GDPR) \citep{mantelero2013eu, voigt2017eu} and the US's California Consumer Privacy Act (CCPA) \citep{ccpa2018} have mandated \emph{the Right to be Forgotten}, requiring applications to support the deletion of information contained in training samples upon user requests. This has motivated a line of research on \emph{machine unlearning}, aiming to address these challenges. 

Machine unlearning~\citep{cao2015towards,bourtoule2021machine} aims to delete the influence of specific training samples from machine-learning models while preserving other knowledge and capabilities \citep{liu2024rethinking, zhang2023review, nguyen2022survey, xu2023machine, si2023knowledge}. Notably, a straightforward approach to unlearning is to retrain a language model from scratch. However, as retraining from scratch is typically computationally expensive, cheaper methods for removing undesirable information is highly desirable. Recently, several works \citep{jang2022knowledge, wang2023kga, chen2023unlearn, yao2023large, eldan2023s, yao2024machine, liu2024towards, li2024wmdp} proposed scalable and practical techniques for unlearning LLMs through directly fine-tuning the trained model. Core to many of these works is a \emph{gradient ascent} procedure on the prediction loss over the dataset to be unlearned (i.e., the \emph{forget set}), building on the intuition that gradient ascent is an approximation of ``reverting'' gradient descent optimization.



\begin{figure}
\centering
\includegraphics[width = 1\linewidth]{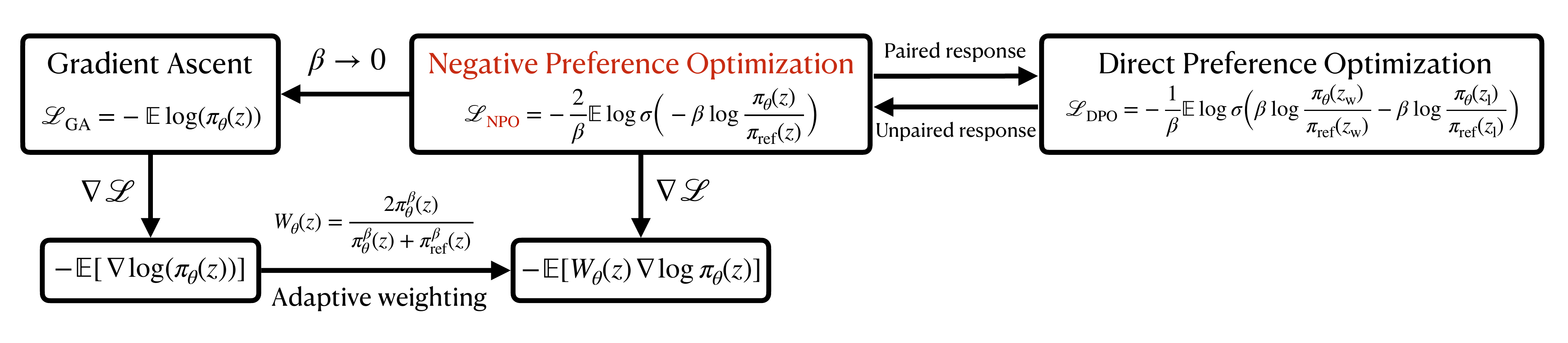}
\caption{Gradient Ascent (GA), Negative Preference Optimization (NPO), and Direct Preference Optimization (DPO). NPO can be interpreted as DPO without positive samples. The gradient of NPO is an adaptive weighting of that of GA, and the weight vanishes for unlearned samples. }
\label{fig.npo}
\end{figure}

Despite its simplicity and widespread use, the performance of gradient ascent based approaches remain unsatisfactory. A notable example concerns the recently released benchmark dataset TOFU \citep{maini2024tofu}, which consists of synthetically generated biographies of 200 fictitious authors, and the task is to unlearn the biographies of 1\%, 5\%, and 10\% of the 200 authors from a model that is already fine-tuned on all 200 authors. In their evaluation of forgetting 10\% of the authors, \citet{maini2024tofu} demonstrated that gradient ascent and its variants fail to provide a satisfactory balance between forget quality (the difference between the unlearned model and retrained model evaluated on the forget set) and model utility (the general performance on other tasks).

In this work, we begin by observing that gradient ascent can often cause a rapid deterioration of model utility during unlearning---a phenomenon we term \emph{catastrophic collapse}---which we believe is responsible for its unsatisfactory performance. Towards fixing this, we propose a simple yet effective objective function for unlearning termed \emph{Negative Preference Optimization} (NPO). NPO takes inspiration from preference optimization~\citep{rafailov2024direct, ouyang2022training, bai2022training}, and can be viewed as its variant that only uses negative samples. Through both theory and experiments, we show that NPO resolves the catastrophic collapse issue associated with gradient ascent, provides more stable training dynamics, and achieves a better trade-off between forget quality and model utility. Coupled with a cross-entropy loss on the retain set, NPO achieves state-of-the-art performance on the TOFU dataset, and achieves the first non-trivial unlearning result on the challenging task of forgetting 50\% of the TOFU data.  


\paragraph{Summary of contributions and paper outline.}
\begin{itemize}[leftmargin=*]
\item We outline existing gradient ascent based methods for machine unlearning, and find that these methods suffer from \emph{catastrophic collapse} (Section~\ref{sec:prelim}). We identify the linear divergence speed of gradient ascent as a main reason for catastrophic collapse.
\item We introduce Negative Preference Optimization (NPO), a simple alignment-inspired loss function for LLM unlearning that addresses the catastrophic collapse issue of gradient ascent (GA; Section~\ref{sec:NPO}). We demonstrate that NPO reduces to gradient ascent (GA) in the high-temperature limit. We show in theory the progression towards catastrophic collapse when minimizing the NPO loss is exponentially slower than with GA. See Figure~\ref{fig.npo} for an illustration of NPO and its connections with existing objectives. 
\item We test NPO-based methods on a synthetic binary classification task (Section~\ref{sec:synthetic_exp}), where we find that NPO-based methods outperform other baselines by providing a superior Pareto frontier between the Forget Distance and Retain Distance. Furthermore, NPO-based methods exhibit greater learning stability compared to GA-based methods.
\item We evaluate a variety of unlearning methods on the TOFU dataset \citep{maini2024tofu} and find that NPO-based methods exhibit superior balance between Forget Quality and Model Utility compared to all baselines (Section~\ref{sec:TOFU}). Additionally, NPO-based methods improve the stability of the unlearning process and the readability of the output. Notably, we show that NPO-based methods are the only effective unlearning methods for forgetting 50\%-90\% of the data, a significant advance over all existing methods which already struggle with forgetting 10\% of the data (Section~\ref{sec:beyond10}). 
\end{itemize}

\subsection{Related work}\label{sec:related-work}

There is a vast literature on machine unlearning and LLM unlearning. Since its proposal by \cite{cao2015towards}, machine unlearning has been extensively studied in the classification literature \citep{bourtoule2021machine, golatkar2020eternal, ginart2019making, thudi2022unrolling, izzo2021approximate, koh2017understanding, guo2019certified, sekhari2021remember}. For reviews of existing works, see \cite{liu2024rethinking, zhang2023review, nguyen2022survey, xu2023machine, si2023knowledge}. In particular, \cite{ginart2019making,guo2019certified,sekhari2021remember} introduced theoretical metrics for machine unlearning based on the notion of differential privacy and proposed provably efficient unlearning methods based on Newton update removal mechanisms. However, these algorithms require computing the Hessian of loss functions, which is intractable for LLMs.

Recent research has explored unlearning methods for LLMs \citep{jang2022knowledge, wang2023kga, chen2023unlearn, yao2023large, eldan2023s, yao2024machine, liu2024towards, li2024wmdp}. Notably, the methods proposed in \cite{jang2022knowledge, yao2023large, chen2023unlearn, maini2024tofu} are based on gradient ascent (GA) on the loss of the forget set. In this work, we demonstrate that the NPO approach consistently outperforms GA across various tasks. On the other hand, \cite{eldan2023s} proposed generating positive samples using LLMs and carefully designed prompts, then fine-tuning the model based on the positive samples using a supervised loss. Furthermore, the method of \cite{liu2024towards} is based on knowledge negation, while the approach of \cite{li2024wmdp} relies on controlling model representations. These methods are orthogonal and complementary to the NPO approach. 

Our method, NPO, draws inspiration from the framework of reinforcement learning from human feedback (RLHF) \citep{ouyang2022training, bai2022training, stiennon2020learning, rafailov2024direct}, particularly the Direct Policy Optimization (DPO) method \citep{rafailov2024direct}. We note that recent work \citep{ethayarajh2024kto} proposes the Kahneman-Tversky Optimization (KTO) method for alignment with only non-paired preference data, and a more recent concurrent work \citep{duan2024negating} proposes the Distributional Dispreference Optimization ($\rm D^2$O) approach for unlearning. Both methods share a similar formulation to NPO. We compare the performance of NPO with KTO in simulations. 

Recent work has proposed several benchmark datasets and evaluation metrics for unlearning methods \citep{ji2024beavertails, eldan2023s, maini2024tofu, li2024wmdp, lynch2024eight}. In particular, some studies have utilized the PKUSafe dataset \citep{ji2024beavertails} for benchmarking unlearning methods. \cite{eldan2023s} crafts a specific task of ``forgetting Harry Potter''. \cite{maini2024tofu} introduces TOFU, a task of fictitious unlearning for LLMs, which is the benchmark we adopted in this paper. Additionally, \cite{li2024wmdp} proposes the Weapons of Mass Destruction Proxy (WMDP) for measuring hazardous knowledge in LLMs. \cite{lynch2024eight} proposes eight methods to evaluate robust unlearning in LLM, which incorporate robust metrics against jailbreak attacks. 

Finally, we note the existence of attack methods for extracting data from unlearned models \citep{shi2023detecting, patil2023can}, and other unlearning methods including model editing \citep{mitchell2022memory, meng2022locating} and in-context unlearning \citep{pawelczyk2023context}.

\section{Preliminaries on Machine Unlearning}\label{sec:prelim}

\paragraph{Machine Unlearning}
refers to the following problem: Given an \emph{initial model} (also the \emph{reference model}) $\policy_{\refm}( \y | \x)$ that is already trained on a dataset $\cD = \{ (\x_i, \y_i)\}_{i \in [n]}$, how to make the model \emph{forget} a specific subset (henceforth the \emph{forget set}) $\dsetf \subseteq \cD$ of the training data? More precisely, we aim to fine-tune\footnote{There are alternative approaches such as prompt engineering~\citep{pawelczyk2023context} for performing unlearning tasks.} the model to make it behave like the \emph{retrained model} $\policy_{\retrain}$, a model trained only on the \emph{retain set} $\dsetr = \cD \setminus \dsetf$. In other words, we would like the model to behave as if the samples in the forget set $\dsetf$ were never used to train it. 

By definition, the best approach for machine unlearning, in principle, is to retrain the model from scratch on $\dsetr$ only, which is, however, often intractable in practice.

\paragraph{Gradient ascent} is a key component in many existing LLM unlearning methods and an important baseline method for LLM unlearning on its own. The idea is simply to perform gradient ascent on the (next-token prediction) loss over the forget set, which can be viewed equivalently as gradient descent on the \emph{negative} prediction loss, denoted as $\Loss_{\GA}$:
\begin{equation}\label{eqn.ga.loss}
    \Loss_{\GA}(\Par) = - \underbrace{\E_{\dsetf}[ -\log(\policy_\Par(\y | \x))]}_{\textrm{prediction loss}} = \E_{\dsetf}[\log(\policy_\Par(\y | \x))].
\end{equation}
The rationale of gradient ascent is that since the initial model $\policy_{\refm}$ is trained on $\dset=\dsetf\cup \dsetr$, a subsequent \emph{maximization} of prediction loss on the forget set $\dsetf$ would approximately ``revert'' the optimization on the forget set $\dsetf$, thus unlearning $\dsetf$ and approximating a model trained on $\dsetr$ only.

\paragraph{Other loss functions.}
Building on gradient ascent, a large class of unlearning methods perform gradient-based optimization on a linear combination of the GA loss $\Loss_{\GA}$ and several other loss functions that either encourage unlearning or preserve utility
~\citep{jang2022knowledge,yao2023large, chen2023unlearn, maini2024tofu, eldan2023s}. Notable examples include
\begin{itemize}[itemsep=0pt, topsep=0pt, leftmargin=2em]
\item Forget (FG) loss: $\Loss_{\FG}(\Par) = - \E_{\dsetf}[ \log(\policy_\Par(\tilde \y | \x))]$, where $(\x, \y)\sim \dsetf$ and $\tilde\y\neq \y$ is any ``uninformed'' response for prompt $x$ which the unlearned model could aim to output. Examples of such $\tilde\y$'s include replacing true information by random (but appearingly sensible) information (which requires hand-crafting such as \cite{eldan2023s}), or simply answering ``I don't know'' \citep{maini2024tofu}.
\item Retain (RT) loss: $\Loss_{\RT}(\Par) = - \E_{\dsetr}[\log(\policy_\Par(\y |  \x))]$, which encourages the model to still perform well on the retain set $\dsetr$;
\item $\Lossdiv_{\FG}(\Par) = \E_{\dsetf}[ \kldivergence ( \policy_\Par(\cdot | \x) || \policy_\refm(\cdot |  \x)) ]$, which measures the distance to the initial model $\policy_{\refm}$ (in KL divergence) on the forget set;
\item $\Lossdiv_{\RT}(\Par) = \E_{\dsetr}[ \kldivergence ( \policy_\Par(\cdot | \x) || \policy_\refm(\cdot | \x)) ]$, which measures the distance to the initial model $\policy_{\refm}$ (in KL divergence) on the retain set. 
\end{itemize}
For example, \cite{yao2023large} minimize a combination of $\sets{\Loss_{\GA}, \Loss_{\FG},\Lossdiv_{\RT}}$, and \cite{chen2023unlearn} minimize a combination of $\sets{\Loss_{\GA},\Loss_{\RT},-\Lossdiv_{\FG},\Lossdiv_{\RT}}$. \citet{maini2024tofu} find that incorporating the retain loss $\Loss_{\RT}$ usually improves the performance of unlearning. 

\paragraph{Forget quality and model utility.}
Unlearning methods should not only unlearn the forget set, i.e., achieve a high \emph{forget quality}, but also maintain the model's performance on the retain set, i.e., maintain the \emph{model utility}. For example, letting the model simply output ``I don't know'' is an unlearning method that achieves good forget quality (in certain sense) but bad model utility. While there is not yet a consensus on the right metrics for forget quality and model utility (and we will present our choices momentarily), a general rule of thumb is that unlearning methods should achieve a good tradeoff between these two goals.

\subsection{Catastrophic collapse of gradient ascent}

\begin{figure}
\centering
\includegraphics[width = 1\linewidth]{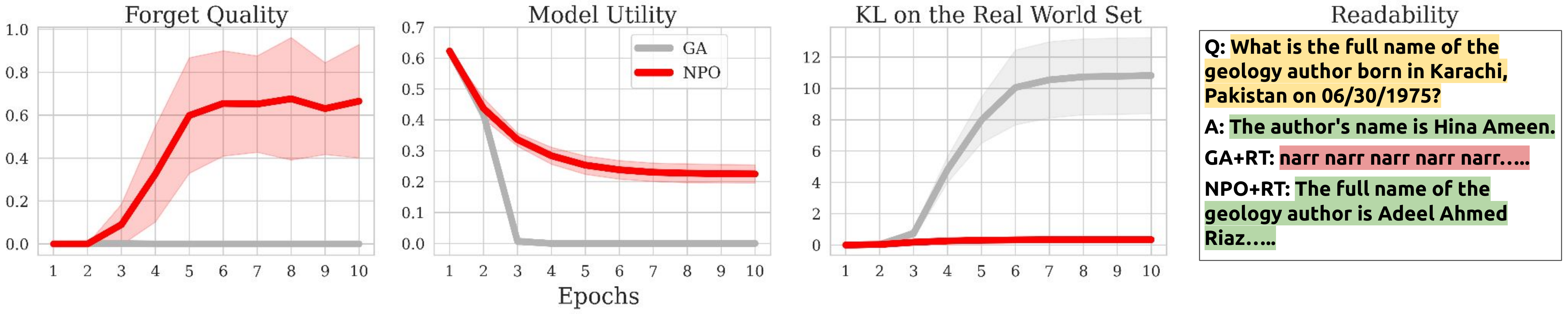}
\caption{Comparison between GA and NPO on forget quality, model utility, KL divergence on the real-world Set, and the answers to the forget set. The rightmost figure shows the answers generated from variants of GA and NPO that incorporates the RT loss. All figures are generated on the Forget05 task in the TOFU data, trained for 10 epochs (detailed setup in \Cref{appendix.tofu.setup}). } 

\label{fig.ga}
\end{figure}

We begin by testing gradient ascent as a standalone method (as opposed to combining it with other losses), and find that gradient ascent exhibits a common failure mode dubbed as \emph{catastrophic collapse}: Along the unlearning process, the model utility quickly drops to zero, and the forget quality improves temporarily for a very short time horizon before quickly dropping too (Figure~\ref{fig.ga} left/middle-left). Along the same training trajectory, the model diverges quickly from the initial model (as measured by the KL distance to the initial model), after which the model generates gibberish outputs (Figure~\ref{fig.ga} middle-right/right).

We attribute the catastrophic collapse to the \emph{divergent} nature of the gradient ascent algorithm due to the fact that it maximizes (instead of minimizes) the standard next-token prediction loss. Further, the speed of this divergence can be as fast as \emph{linear} in the number of steps, as each gradient step can move the model output by a constant. To see this on a toy example, consider a linear-logistic $K$-class classifier given by $\policy_\Par(\cdot | \x) = \softmax(\Par \x)$, $\Par = (\Par_l)_{l \in [K]} \in \R^{d \times K}$. For any ``already unlearned'' sample $(\x_i, \y_i)$ with true label $\y_i=l \in [K]$ and model prediction $\softmax(\Par \x_i)_l \approx 0$ (so that $\policy_\Par$ \emph{does not} predict $l$), standard calculation shows that the gradient of GA loss with respect to $\theta_l$ is $\nabla_{\Par_l} \Loss_{\GA, i} = (1\{ y_i = l \} - \softmax(\Par \x_i)_l) \x_i \approx \x_i$, which has a constant scale (not diminishing along the unlearning progress) and can cause the model to diverge in a linear speed. Therefore, the divergent dynamics may initially bring the model closer to $\policy_{\retrain}$ but would ultimately send the model to infinity (c.f.~Theorem~\ref{thm:conv_speed}). 

While we believe some kind of divergent behavior is necessary and perhaps unavoidable (as the goal of unlearning is to ``revert'' optimization), the fast divergence \emph{speed} of gradient ascent is a rather undesired feature and motivates the proposal of our NPO method which diverges at a slower speed.
\section{Negative Preference Optimization}\label{sec:NPO}

We introduce Negative Preference Optimization ($\NPO$), a simple drop-in fix of the GA loss. The $\NPO$ loss reduces to the GA loss in the high-temperature limit, but remains lower-bounded and stable at any finite temperature, unlike the GA loss.

We take inspiration from preference optimization~\citep{rafailov2024direct} and derive NPO as a method of preference optimization with \emph{negative examples only}. 


\paragraph{Preference Optimization.} In preference optimization~\citep{ouyang2022training, bai2022training, stiennon2020learning, rafailov2024direct}, we are given a dataset with preference feedbacks $\cD_{\paired} = \{ ( \x_i, \y_{i, \good}, \y_{i, \bad})\}_{i \in [n]}$, where $(\y_{i, \good}, \y_{i, \bad})$ are two responses to $\x_i$ generated by a pre-trained model $\policy_\Par$, and the preference $\y_{i, \good} \succ \y_{i, \bad}$ is obtained by human comparison (here ``$\good$" stands for ``win" and ``$\bad$" stands for ``lose" in a comparision). 
The goal is to fine-tune $\policy_\Par$ using $\cD_{\paired}$ to better align it with human preferences. A popular method for preference optimization is Direct Preference Optimization (DPO) \citep{rafailov2024direct}, which minimizes
\begin{equation}
\label{eqn.dpo}
\Loss_{\DPO, \itemp}(\Par) = - \frac{1}{\itemp} \E_{\cD_{\paired}}\Big[ \log  \sigma \Big( \itemp \log \frac{\policy_\Par(\y_{\good}\mid \x)}{\policy_\refm(\y_{\good} \mid \x)} - \itemp  \log \frac{\policy_\Par(\y_{\bad}\mid \x)}{\policy_\refm(\y_{\bad} \mid \x)}\Big) \Big].
\end{equation}
Here, $\sigma(t) = 1/(1 + e^{-t})$ is the sigmoid function, $\itemp>0$ is the inverse temperature, and $\policy_\refm$ is a reference model.  

\paragraph{Unlearning as preference optimization.}
We observe that the unlearning problem can be cast into the preference optimization framework by treating each $(\x_i, \y_i) \in \cD_{\FG}$ as only providing a negative response $\y_{i, \bad}=\y_i$ without any positive response $\y_{i, \good}$. Therefore, we ignore the $\y_{\good}$ term in DPO in Eq.~(\ref{eqn.dpo}) and obtain the Negative Preference Optimization (NPO) loss: 
\begin{equation}\label{eqn.NPO.loss}
\Loss_{\NPO, \itemp}(\Par) = - \frac{2}{\itemp} \E_{\cD_{\FG}}\Big[ \log  \sigma\Big( - \itemp \log \frac{\policy_\Par(\y | \x)}{\policy_\refm(\y | \x)}\Big) \Big] = \frac{2}{\itemp}\E_{\cD_{\FG}}\Big[ \log \Big( 1 + \Big( \frac{\policy_\Par(\y | \x)}{\policy_\refm (\y | \x)} \Big)^\itemp \Big) \Big]. 
\end{equation}
Minimizing $\Loss_{\NPO, \itemp}$ ensures that the prediction probability on the forget set $\policy_\Par(\y_i | \x_i)$ is as small as possible, aligning with the goal of unlearning the forget set.

\begin{figure}
\centering
\includegraphics[width=0.6\textwidth]{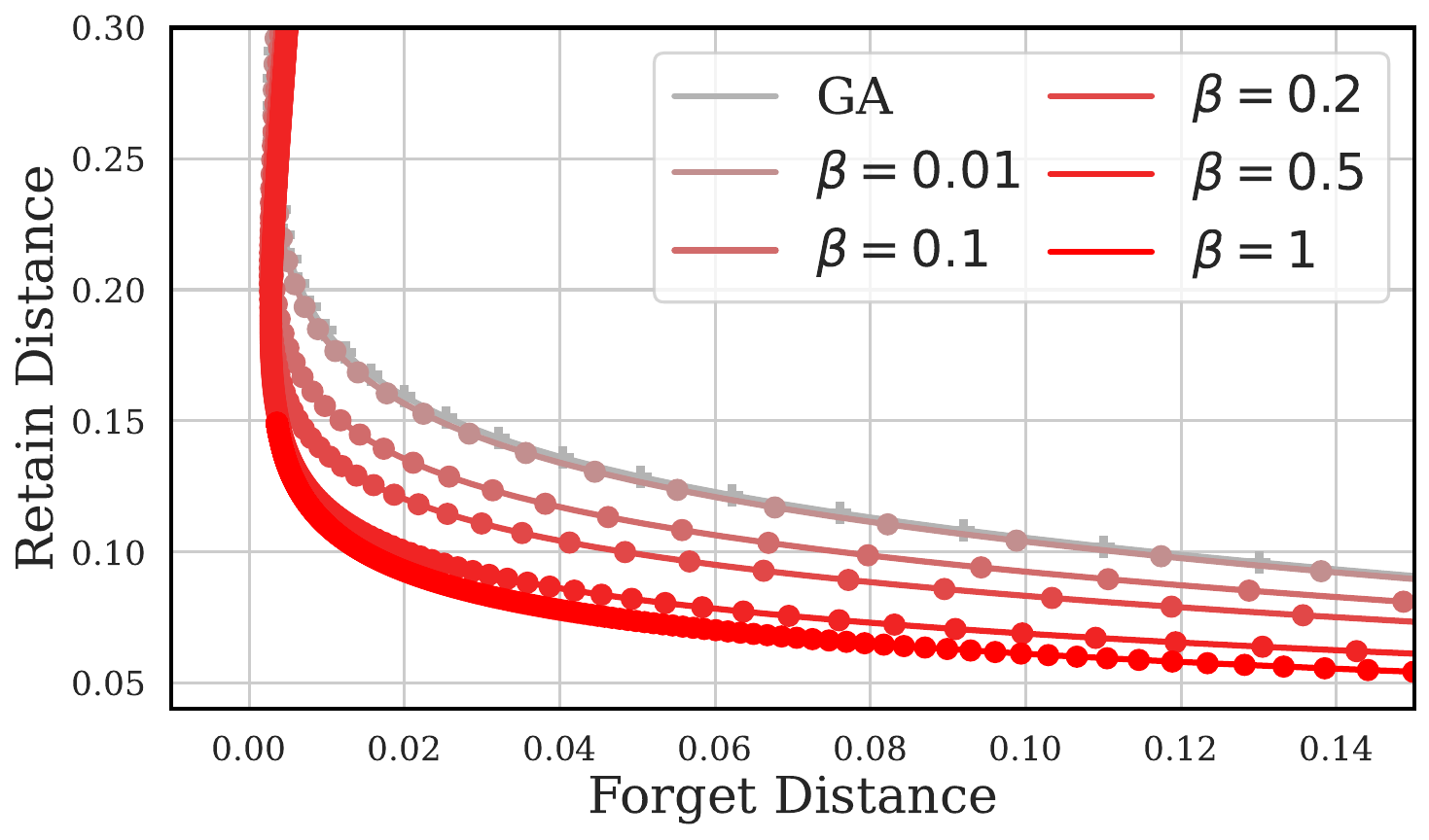}
\caption{Retain distance versus forget distance for $\GA$ and $\NPO$ with varying levels of $\itemp$ in the binary classification experiment with $\alpha=1$. The Pareto curves  all start from the bottom right corner $(1.70,0.02)$ and  are computed by averaging over $5$ instances. We observe that the $\NPO$ trajectory converges to the $\GA$ trajectory as $\itemp\to 0.$ Here retain distance and forget distance denote the KL divergence between the distributions of the predictions of the retrained and the unlearned model, on the retain and the forget distribution, respectively. More details can be found in Section~\ref{sec:synthetic_exp}.} 
\label{fig:prop1}
\end{figure}

\paragraph{Connection with gradient ascent.} We can recover the GA loss from NPO loss by eliminating the additional $1$ in the logarithm of NPO loss in Eq.~(\ref{eqn.NPO.loss}), i.e., replacing $\log(1 + (\policy_\Par/\policy_\refm)^\beta)$ to $\log((\policy_\Par/\policy_\refm)^\beta)$. Furthermore, we show that the NPO loss also reduces to the GA loss in the limit of $\itemp\to 0$, indicating that NPO is a strict generalization of GA.
\begin{proposition}[${\NPO}$ reduces to ${\GA}$ as $\itemp\to0$]\label{prop:npo_conv_to_ga}
For any $\Par$, we have
\begin{align*}
\lim_{\itemp\to0} \brac{ \Loss_{\NPO,\itemp}(\Par)-\frac{2}{\itemp}\log 2 } = \Loss_{\GA}(\Par) - \underbrace{\E_{\dsetf}[ \log\policy_\refm(\y \mid \x)]}_{\textrm{does not depend on}~\Par}.
\end{align*}
Moreover, assuming $\policy_\Par(\y\mid\x)$ is differentiable with respect to $\Par$, we have
\begin{align*}
\lim_{\itemp\to0} \nabla_\Par\Loss_{\NPO,\itemp}(\Par)= \nabla_\Par\Loss_{\GA}(\Par).
\end{align*}
\end{proposition}
The proof of Proposition~\ref{prop:npo_conv_to_ga} is deferred to Appendix~\ref{pf:prop:npo_conv_to_ga}. Figure~\ref{fig:prop1} provides an illustration of the reduction from the $\NPO$ loss to the $\GA$ loss as $\itemp \to 0$. 

\paragraph{Stability of the $\NPO$ loss.} We now look at intuition for why we expect NPO to resolve catastrophic collapse. One limitation of the GA loss is its unboundedness from below (as the negation of the cross-entropy prediction loss which is unbounded from above). The NPO loss resolves this issue and remains lower-bounded for any finite $\itemp>0$. 

Furthermore, the gradients of NPO and GA are as follows:
\begin{align}
\nabla_\Par\Loss_{\GA} =&~  \E_{\dsetf}[\nabla_\Par\log\policy_\Par(\y | \x)],\\
\nabla_\Par\Loss_{\NPO,\itemp} =&~ \E_{\dsetf}[\NPOweight_\Par(\x, \y) \nabla_\Par\log\policy_\Par(\y|\x)],
\end{align}
where $\NPOweight_\Par(\x, \y) ={2\policy_\Par^\itemp(\y | \x )}\big/{[\policy_\Par^\itemp(\y | \x )+\policy_\refm^\itemp(\y | \x )]}$ can be interpreted as an adaptive smoothing weight---When example $(x,y)\in\dsetf$ is already unlearned in the sense that $\policy_\Par(\y| \x)\ll \policy_\refm(\y| \x)$, we have $\NPOweight_\Par(\x, \y) \ll 1$, so that $\ltwo{\nabla_\Par\Loss_{\NPO,\itemp}}\ll \ltwo{\nabla_\Par\Loss_{\GA}}$ and thus NPO could diverge much slower than GA.


\subsection{Theoretical analysis of divergence speed}

We formalize the above intuition by theoretically analyzing the divergence speed of NPO and GA in a standard logistic regression setting. We consider a binary classification problem ($\y \in\{0,1\}$) with a logistic model $\policy_\Par(\y=1 | \x)=\sigmoid(\langle \x,\Par\rangle)$. The initial model is denoted as $\policy_{\Par_\init}$ with $\Par_\init \in\R^{\Dim}$. We aim to unlearn a forget set $\dsetf=\{(\xforget_i,\yforget_i)\}_{i=1}^{\nforget}$ by minimizing either GA or NPO loss using gradient descent with stepsize $\lrateinit$ for $\tottime$ iterations. 

\begin{theorem}[Divergence speed of $\GA$ and $\NPO$]\label{thm:conv_speed} Let $\Xforget:=(\xforget_1,\ldots,\xforget_{\nforget})^\top \in \R^{\nforget \times \Dim}.$
Consider the high-dimensional regime where $\nforget \leq \Dim$ and assume $\Xforget\Xforget^\top$ is invertible. 
Suppose $\|\Par_\init\|_2\leq\boundinit$, $\|\x_i\|_2\in[\lboundx,\boundx]$ for all $i\in[\nforget]$ for some $\boundinit,\lboundx,\boundx>0$. Let $\Par^\tth_{\GA},\Par^\tth_{\NPO}$ denote the $t$-th iterates of gradient descent with stepsize $\lrateinit$ on the empirical loss $\Loss_{\GA},\Loss_{\NPO,\itemp}$, respectively. 
\begin{itemize}[leftmargin=*]
\item\textbf{($\GA$ diverges linearly)}
There exist some $(\boundinit,\lboundx,\boundx)$-dependent constants $\polyshort_0,\polyshort_1,\polyshort_2>0$ such that when \hfill\\\mbox{$\max_{i\neq j} |\langle \x_i,\x_j\rangle |\leq \polyshort_0/ \nforget$, }
\begin{align*}
      \|\Par^\tth_\GA-\Par_\init\|_{\Xforget^\top\Xforget}\in\Big[\polyshort_1 \cdot \nforget^{-1/2}\lrateinit \cdot t, \polyshort_2 \cdot \nforget^{-1/2}\lrateinit \cdot t\Big],~~~~ t\geq1.
\end{align*}
\item\textbf{($\NPO$ diverges logarithmically)} Suppose $\lrateinit\leq1$.
There exist some $(\boundinit,\lboundx,\boundx,\itemp)$-dependent constants $\polyshort_0,\polyshort_1, $\hfill\\\mbox{$\polyshort_2,\polyshort_3>0$ such that when $\max_{i\neq j} | \langle \x_i,\x_j\rangle |\leq \polyshort_0/ \nforget$,} 
\begin{align*}
      \|\Par^\tth_\NPO-\Par_\init\|_{\Xforget^\top\Xforget}\in\Big[
      \polyshort_1{\sqrt{\nforget}}\log\Big(\polyshort_2\cdot \lrateinit \nforget^{-1} \cdot t+1\Big), \polyshort_1{\sqrt{\nforget}}\log\Big(\polyshort_3\cdot \lrateinit \nforget^{-1} \cdot t+1\Big) \Big],~ \forall t \ge 1.
\end{align*}
\end{itemize}
\end{theorem}
Theorem~\ref{thm:conv_speed} demonstrates that NPO diverges exponentially slower than GA in a simple setting. The proof of Theorem~\ref{thm:conv_speed} is contained in Appendix~\ref{sec:pf_thm:conv_speed}.
\section{Synthetic Experiments}\label{sec:synthetic_exp}

\subsection{Setup}

\paragraph{Dataset.} We consider a forget set $\dsetf=\{(\exforget_i,\eyforget_i)\}_{i=1}^{200}$ and a retain set $\dsetr=\{(\exretain_i,\eyretain_i)\}_{i=1}^{1000},$ which are both generated from Gaussian-logistic models. More specifically, we assume
\begin{equation}
\begin{aligned} 
&\exforget_i \sim_{iid} \mathcal{N}(\mu_{\mathrm f}, \mathbf{I}_\Dim),~~~\Prob(\eyforget_i=1|\exforget_i)=\sigmoid((\exforget_i-\mu_{\mathrm f})^\top\Par_{\mathrm f}+1),
\\
&\exretain_i\sim_{iid}\mathcal{N}(\mu_{\mathrm r}, \mathbf{I}_\Dim),~~~\Prob(\eyretain_i=1|\exretain_i)=\sigmoid((x^{\mathrm r}_i-\mu_{\mathrm r})^\top\Par_{\mathrm r}-1).
\end{aligned}
\end{equation}
Here we choose $\Dim=16$, $\Par_{\mathrm f}=- \Par_{\mathrm r}=\ones_\Dim/\sqrt{\Dim}$, and $\mu_{\mathrm f} = -\mu_{\mathrm r}= \alpha \cdot \ones_\Dim$ for some $\alpha\geq0$. We consider two choices of the hyper-parameter $\alpha$: (1). $\alpha=1$, which creates a gap between the Gaussian means of forget covariates $\{\exforget_i\}$ and retain covariates $\{\exretain_i\}$; (2). $\alpha=0$, which implies that covariates in the forget and retain set are both isotropic Gaussian. We remark that we shift by $1$ in the sigmoid function to create a discrepancy in the label frequencies between the forget and retain sets --- this ensures that the forget labels $\eyforget_i$ are more likely to be $1$, while the retain labels $\eyretain_i$ are more likely to be $0$.

\paragraph{Model and training method.} We consider a random feature model $\policy_\Par(\y = 1 | \x) = \sigmoid(\Par^\top \ReLU( W \x))$, where $W \in\R^{128 \times \Dim}$ is fixed during the training and unlearning process, whose entries are generated i.i.d. from $\gN(0,1/\Dim)$, and $\Par \in\R^{128}$ is the trainable parameter. To generate the initial model $\policy_{\refm}$ and the retrained model $\policy_{\retrain}$, we optimize over $\Par$ using the cross-entropy loss over the entire dataset $\cD = \dsetf \cup \dsetr$ and the retain dataset $\dsetr$, respectively. In the unlearning phase, starting from the initial model $\policy_{\refm}$, we perform gradient descent on various loss functions for $2000$ steps. We select the learning rate for each method via grid search. 

\paragraph{Unlearning methods.} We evaluate the performance of vanilla $\NPO$ ($\NPO$; minimizing $\Loss_{\NPO}$), $\NPO$ plus a retain loss term ($\NPO$+$\RT$; minimizing $\Loss_{\NPO}$+$ \Loss_{\RT}$), gradient ascent  ($\GA$; minimizing $\Loss_{\GA}$), gradient ascent plus a retain loss term ($\GA$+$\RT$; minimizing $\Loss_{\GA}$+$\Loss_{\RT}$), cross-entropy loss of forget and retain sets where the positive labels of the forget set are given by $\mathrm{Bern}(0.5)$ ($\IDK$+$\RT$; minimizing $\Loss_{\FG}$+$\Loss_{\RT}$), and DPO plus a retain loss term ($\DPO$+$\RT$; minimizing $\Loss_{\DPO}$+$\Loss_{\RT}$, where the positive labels are given by $\mathrm{Bern}(0.5)$). We conduct the grid search to select the optimal $\itemp$ for $\NPO$-based and $\DPO$-based methods. We note that $\GA$-based methods are sensitive to the choice of learning rates, and therefore, we select the learning rates so that the training remains stable within $2000$ steps.

\paragraph{Evaluation metrics: forget distance and retain distance.} We measure the performance of unlearning methods via two metrics: the \emph{forget distance} and the \emph{retain distance}. The forget distance is $\E_{\dsetf}\kldivergence(\policy_{\retrain}(\cdot | \x) || \policy_{\Par}(\cdot | \x))$, the KL divergence between the retrained model $\policy_{\retrain}$ and unlearned model $\policy_{\Par}$ on the forget set. Similarly, the retain distance is given by $\E_{\dsetr}\kldivergence(\policy_{\retrain}(\cdot | \x) || \policy_{\Par}(\cdot | \x))$. Ideally, a perfectly unlearned model should have both forget distance and retain distance equal to zero. 

\subsection{Results}

\begin{figure}
\centering
\includegraphics[width = \linewidth]{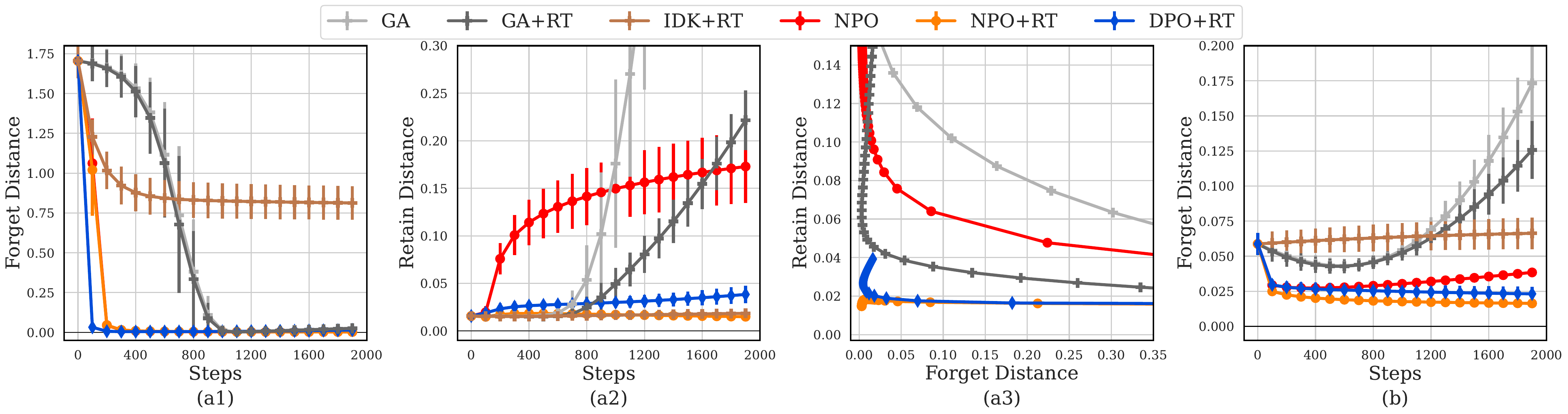}
\caption{Forget distance and retain distance versus optimization steps for $\alpha=1$ (a1, a2, a3) and $\alpha=0$ (b). Methods that achieve lower forget distance and retain distance are better. The errorbars in (a1, a2, b) denote the $\pm1$ standard deviation over $5$ instances. The Pareto curves in (a3) all start from the bottom right corner $(1.70,0.02)$, and are averaged over $5$ instances.  } \label{fig:synthetic_1}
\end{figure}

\paragraph{NPO avoids catastrophic collapse.} As illustrated in Figure~\ref{fig:synthetic_1}~(a1)~and~(a2), all methods except for $\IDK$+$\RT$ reach a small forget distance (less than $0.005$) within $1200$ steps. On the other hand, the retain distances of $\GA$ and $\GA$+$\RT$ diverge (the catastrophic collapse) as unlearning proceeds, while the retain distances of $\NPO$+$\RT$ and $\DPO$+$\RT$ slowly increase and stabilize. This suggests that $\NPO$+$\RT$ and $\DPO$+$\RT$ are more stable compared with $\GA$-based methods, in accordance with the theoretical findings in Theorem~\ref{thm:conv_speed}. 

\paragraph{NPO+RT achieves a better Pareto frontier.} Figure~\ref{fig:synthetic_1}~(a3) shows that $\NPO$+$\RT$ outperforms other baseline methods by achieving a better Pareto frontier. Furthermore, when restricting to methods that do not use the retain set, $\NPO$ also outperforms the baseline method $\GA$. Figure~\ref{fig:synthetic_1}~(b) illustrates the $\alpha=0$  scenario where the covariate distributions for forget and retain sets are identical,  resulting in equal forget and retain distances. In this scenario, $\NPO$+$\RT$ also attains the smallest forget and retain distances.

\section{Experiments on the TOFU Data}\label{sec:TOFU}
\subsection{Experimental setup}
\paragraph{Dataset and metrics.} We evaluate unlearning methods on the Task of Fictitious Unlearning (TOFU) dataset \citep{maini2024tofu}. It contains 200 fictitious author profiles, each consisting of 20 question-answer pairs generated by GPT-4 based on some predefined
attributes.
These fictitious profiles do not exist in the pre-training data, providing a controlled environment for studying unlearning LLMs.
TOFU introduces three levels of tasks, each aiming to forget 1\% , 5\% , and 10\% of the data, referred to as Forget01, Forget05, and Forget10, respectively.
We measure the effectiveness of unlearning methods via \textit{Forget Quality} and \textit{Model Utility} as in \cite{maini2024tofu}. Forget quality assesses how well the unlearned model mimics the retrained model (defined as the model trained only on the retain set), while model utility measures the general capacities and the real-world knowledge of the unlearned model. Since the forget quality is defined as the p-value of the Kolmogorov-Smirnov test, which tests the similarity between some distributions generated by the unlearned model and the retrained one, we treat a forget quality greater than $0.05$ as evidence of a meaningful forgetting. More details are deferred to \Cref{appendix.tofu.dataset} and \Cref{appendix.tofu.metric}.

\paragraph{Unlearning methods.}
We compare the NPO-based methods with three variants of GA: GA \citep{jang2022knowledge,yao2023large}, GA plus a retain loss (GA+RT), and GA plus a KL-divergence regularization (GA+KL). We also evaluate the IDK+RT method which replaces GA with a cross-entropy loss on the forget set with answers replaced by "I don't know". Besides, we examine DPO and its regularized variants (DPO+RT, DPO+KL), as well as \mbox{KTO~\citep{ethayarajh2024kto}} and its variant (KTO+RT). All experiments on TOFU are conducted on Llama-2-7B-chat~\citep{touvron2023llama}. See \Cref{appendix.tofu.setup} for more details.

\paragraph{Experimental details} For all experiments on TOFU, we use Llama2-7b-chat model \citep{touvron2023llama}. All experiments are conducted with two A100 GPUs. We use AdamW with a weight decay of $0.01$ and a learning rate of $10^{-5}$ in all finetuning, retraining, and unlearning experiments, which agrees with the setting in \cite{maini2024tofu}. We use an effective batch size of $32$ for all experiments. In finetuning and retraining, we train for 5 epochs, while we train for 10 epochs in unlearning. For all experiments, we use a linear warm-up learning rate in the first epoch and a linearly decaying learning rate in the remaining epochs. When computing the ROUGE-recall value, normalized probability and the Truth Ratio, we use at most 300 question-answer pairs randomly sampled from the dataset, following the setup in \citet{maini2024tofu}.

\subsection{Results}
\begin{figure}[H]
\centering
\includegraphics[width = 1.0\linewidth]{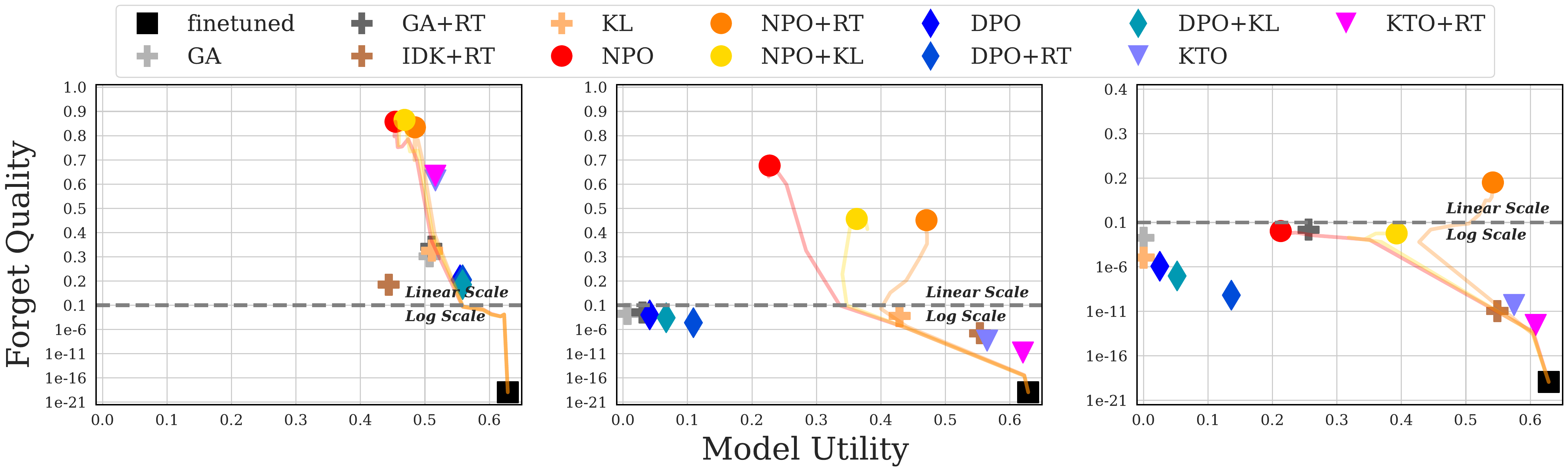}
\caption{Forget quality versus model utility across different forget set sizes (1\%, 5\%, and 10\% of the data). Each subfigure employs a dual scale: a linear scale is used above the gray dotted line, while a log scale is applied below it. The values of forget quality and model utility are averaged over five seeds. Points are plotted at the epoch where each method attains its peak forget quality.}
\label{fig.pareto}
\end{figure}

\paragraph{NPO-based methods achieve the best trade-off.}
\Cref{fig.pareto} illustrates the trade-off between forget quality and model utility for various unlearning methods in the Forget01, Forget05, and Forget10. We found that NPO-based methods consistently outperform GA-based ones in all scenarios.
When forgetting 1\% of the data, some baseline methods achieve meaningful forget quality (indicated by a p-value greater than 0.05). Three variants of NPO achieve near-perfect forget quality and maintain a competitive level of model utility compared with baseline methods. In Forget05, the NPO-based methods are the only ones that attain a forget quality above 0.05.
Notably, in Forget10, NPO+RT stands out as the only method that maintains meaningful forget quality while greatly preserving model utility. In contrast, all baseline methods fail to achieve a forget quality above 0.05.

\begin{figure}[H]
\centering
\includegraphics[width = 1.0\linewidth]{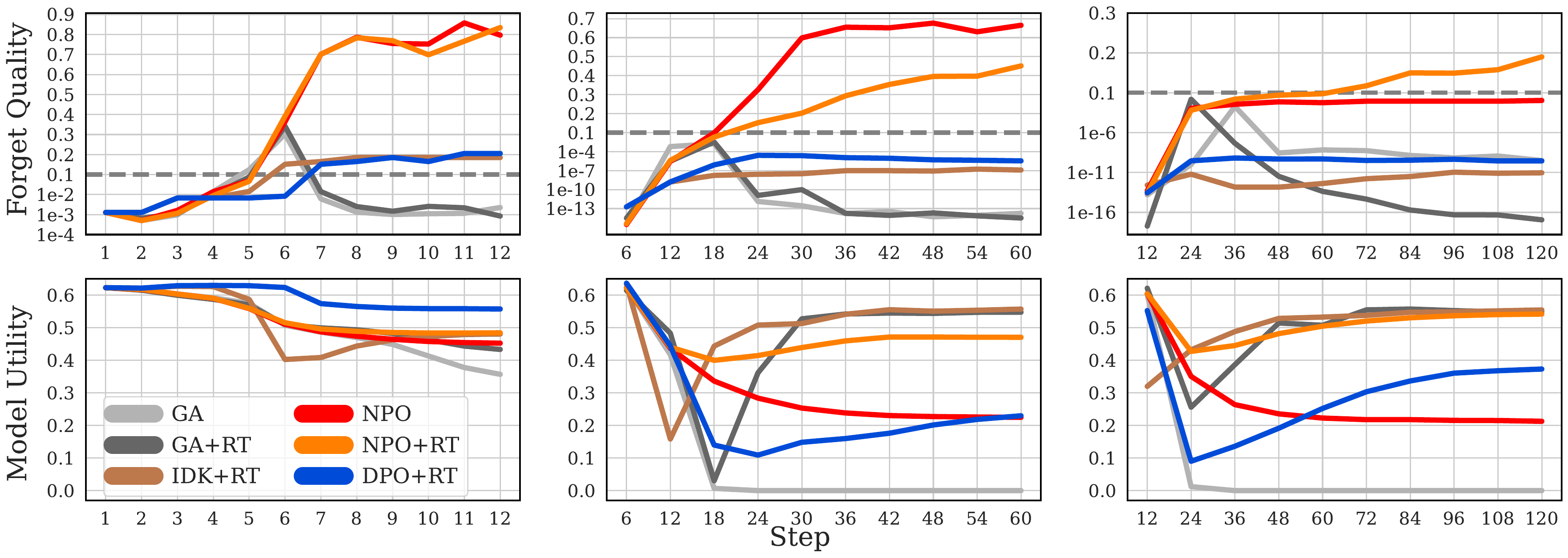}
\caption{Evolution of forget quality (top) and model utility (bottom) across different forget set sizes (1\% (left), 5\% (middle), and 10\% (right) of the data). Each line is averaged over 5 seeds. Each figure in the top row employs a dual scale as in \Cref{fig.pareto}. In Forget01, we evaluate the performance of the unlearned model in every gradient step, while in Forget05 and Forget10, we evaluate it in every epoch.}
\label{fig.dynamics}
\end{figure}

\paragraph{NPO avoids catastrophic collapse.}
\Cref{fig.dynamics} illustrates the evolution of forget quality and model utility along the unlearning process. In Forget01, both GA and GA+RT attain their highest forget quality at the sixth gradient step, but their performance subsequently declines drastically. Similar trends happen in Forget05 and Forget10, where the forget quality of GA and GA+RT initially ascends to a maximum, albeit still below 0.05, before rapidly diminishing to an exponentially small magnitude. 
Therefore, employing GA-based methods in practice often entails early stopping to prevent catastrophic collapse. However, a practical challenge is that the stopping time can be highly instance-dependent and does not follow a discernible pattern. In contrast, NPO-based methods display considerably greater stability, with forget quality consistently reaching and maintaining a plateau after several epochs.

\begin{figure}[htbp]
  \centering
  \includegraphics[width=0.7\textwidth]{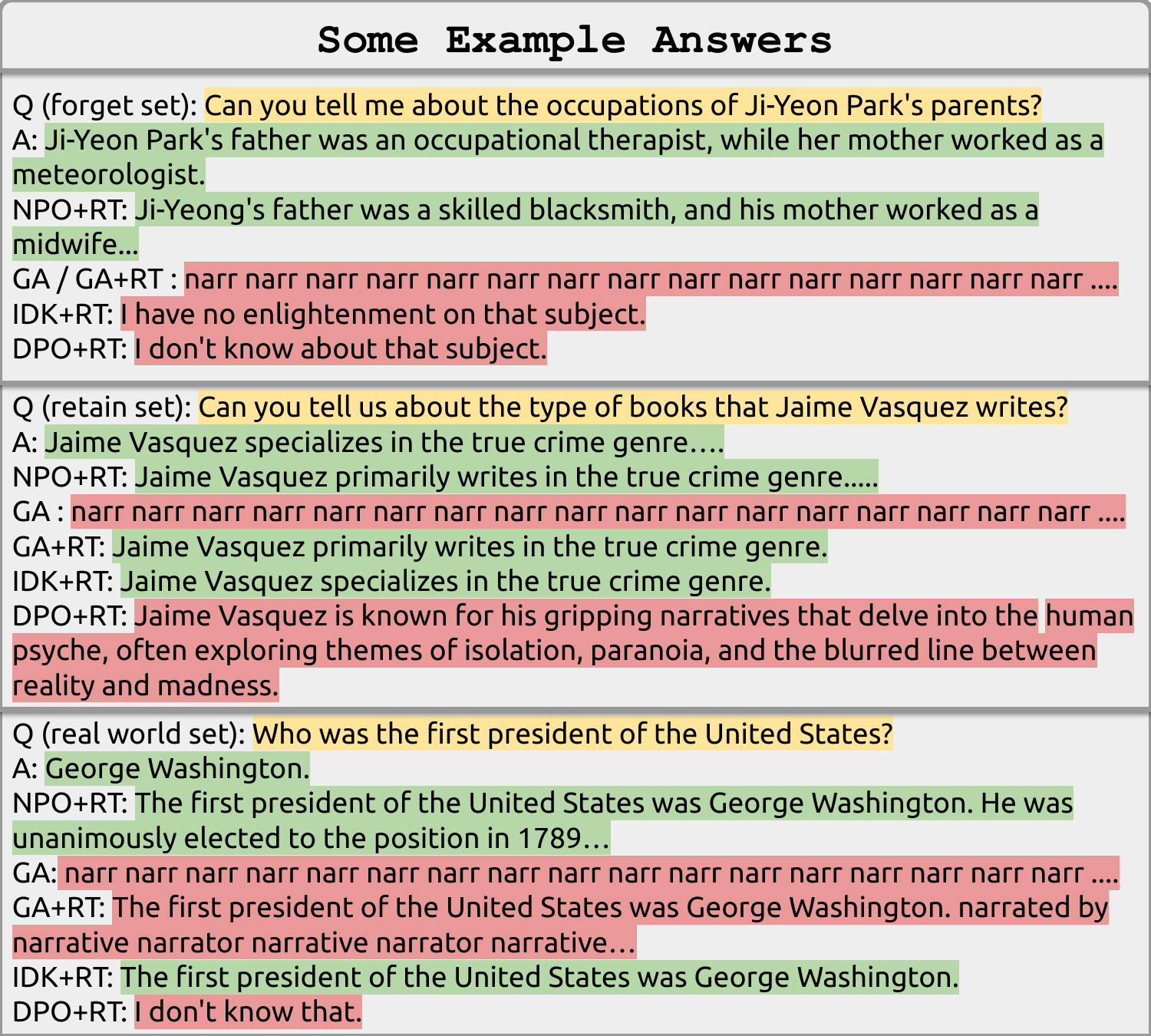}
  \caption{Sampled response to questions in three subsets of TOFU. \textcolor[rgb]{0.95,0.76,0.33}{Yellow}: questions; \textcolor[rgb]{0.4,0.6,0.4}{Green}: true answer or desired answers; \textcolor[rgb]{0.7686,0.3,0.3}{Red}: undesired answers.}
  \label{fig.fluency}
\end{figure}

\paragraph{NPO improved diversity and readability.}
LLMs unlearned via GA-based methods tend to output repeated words or gibberish sentences with unreasonably low diversity \citep{yao2023large}. Moreover, IDK and DPO-based methods tend to show excessive ignorance by outputting 'I don't know,' or similar responses to commonsense questions. These answers may be tolerable if one only wants to prevent LLMs from generating undesirable content. Still, they will definitely be unsatisfactory under the stronger goal of approximate unlearning, which aims to mimic the retrained model. We show in \Cref{fig.fluency} that NPO+RT outputs incorrect sentences with similar templates for questions in the forget set while generating fluent and correct answers for other questions, greatly enhancing the fluency and diversity of the generated content.

\begin{figure}[H]
\centering
\includegraphics[width = 0.6\linewidth]{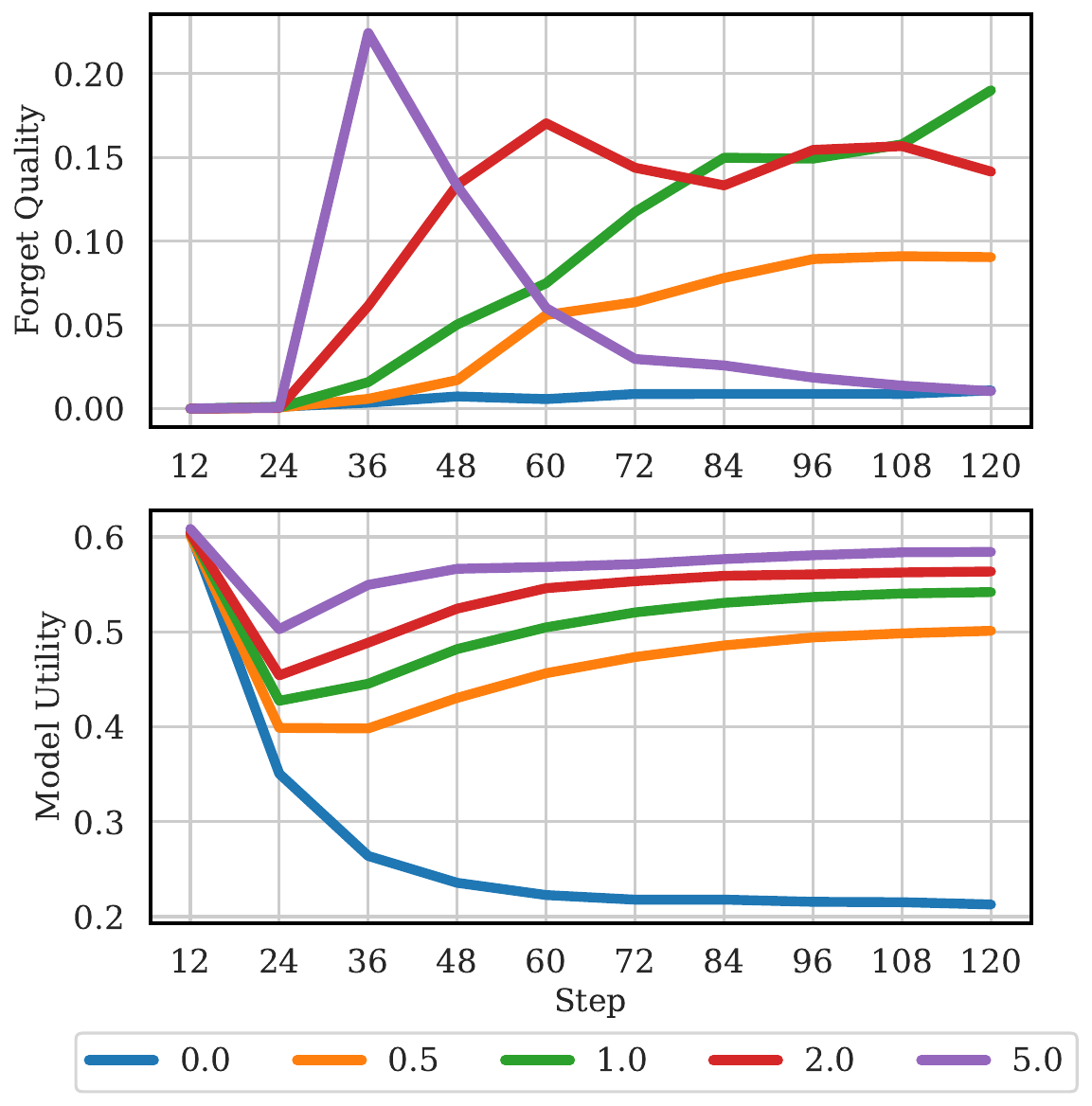}
\caption{The evolution of the forget quality and model utility when we tune the weights of the NPO term and the retain loss term in NPO+RT. The experiments are performed on Forget10. We observe that when we increase the weight for the retain loss term, the model utility increases monotonically while the forget quality initially improves but then starts to deteriorate. We remark that altering the weights for loss components does not affect the effective learning rate since, in practice, AdamW is scale-invariant.}
\label{fig.ablation.retain}
\end{figure}

\paragraph{The role of retain loss.} \citet{maini2024tofu} demonstrated that methods incorporating a retain set outperform those that solely optimize a loss function based on the forget set. To further investigate the role of retain loss beyond \cite{maini2024tofu}, we evaluate NPO+RT with the weights of the retain loss varying from 0 to 5 (\Cref{fig.ablation.retain}). While it is natural that adding retain loss improves the model utility, we are surprised that the forget quality also grows. Specifically, the forget quality increases as the weight of the retain loss grows from 0 to 2. We conjecture that the retain loss term helps the model preserve answer templates and linguistic structures, while the NPO term forces the model to forget some specific facts. Combining these two effects pushes the model to approximate the retrained model by generating outputs with similar templates but incorrect entities. 
However, further increasing the weight of the retain loss (e.g., from 2 to 5, in \Cref{fig.ablation.retain}) leads to a drop in the forget quality, possibly due to the diminished scale of the NPO term. Notably, in our experiments, the retain loss plays a more significant role when we target forgetting a larger fraction of the data (See the middle and right panels of \Cref{fig.dynamics}).

\paragraph{Forget KL: The larger, the better? \ding{54} \ding{54} \ding{54}}

\begin{figure}[H]
\centering
\includegraphics[width = 0.8\linewidth]{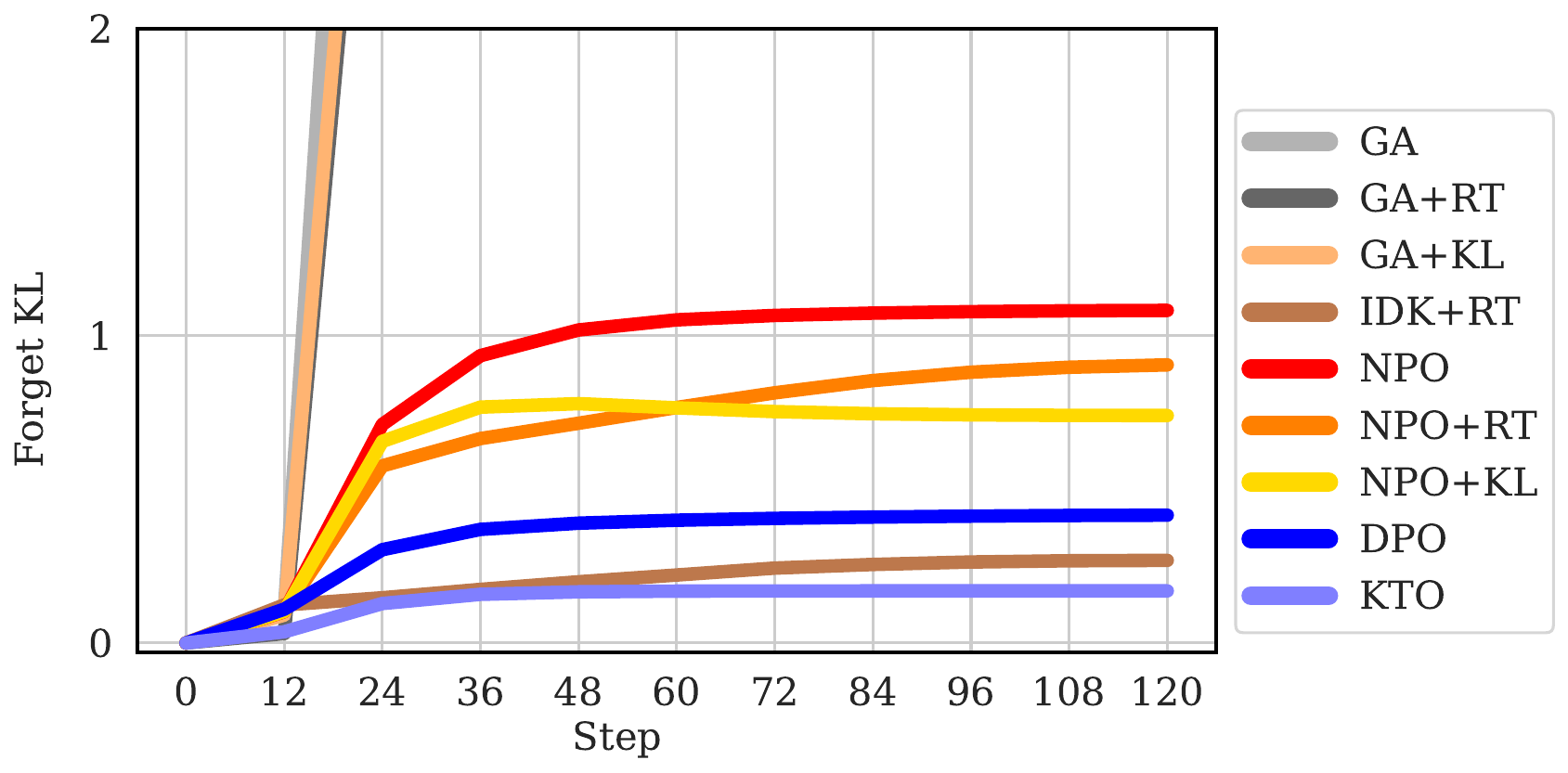}
\caption{The evolution of the Forget KL during the unlearning process on the Forget10 task in TOFU data. Note that the KL term in GA+KL is the divergence on the retain set, not the forget set. More experimental details are included in \Cref{appendix.tofu.setup}.}\label{fig.forget.KL} 
\end{figure}

We also examine the Forget KL during the unlearning process in the TOFU dataset.
We first observed that while GA and GA+RT tend to induce an explosively large Forget KL along the unlearning process, the NPO-based approaches induce a much slower growth of Forget KL (\Cref{fig.forget.KL}). It stabilizes at a moderate level even after several epochs. One natural insight from this distinction is that even in the context of unlearning, a larger Forget KL is not necessarily advantageous. Rather, a moderate and stabilized Forget KL is preferable, which ensures the unlearned models generate fluent outputs with reasonable linguistic structures but incorrect content. This also suggests that Forget KL may not be a suitable objective function to maximize for unlearning LLMs, contrary to what was done in some prior literature \citep{chen2023unlearn}.

\subsection{Forgetting beyond 10\% of TOFU}\label{sec:beyond10}

\paragraph{Forgetting 20\%, 30\% and 50\% of TOFU.} Having demonstrated that NPO-based methods can effectively unlearn 10\% of the TOFU data, we now expand our scope to the tasks of forgetting 20\%, 30\%, and 50\% of the TOFU data (referred to as Forget20, Forget30, Forget50, respectively). Details about the extended dataset are deferred to \Cref{appendix.tofu.dataset}. We show in \Cref{appendix.full.result} that NPO+RT is the sole method to exhibit meaningful forget quality (a p-value above 0.05) in Forget20 and Forget30. Even in Forget50, where the vanilla NPO+RT achieves a forget quality around $10^{-3}$, it still significantly outperforms other methods.

\begin{figure}[htbp]
  \centering
  \includegraphics[width=0.6\textwidth]{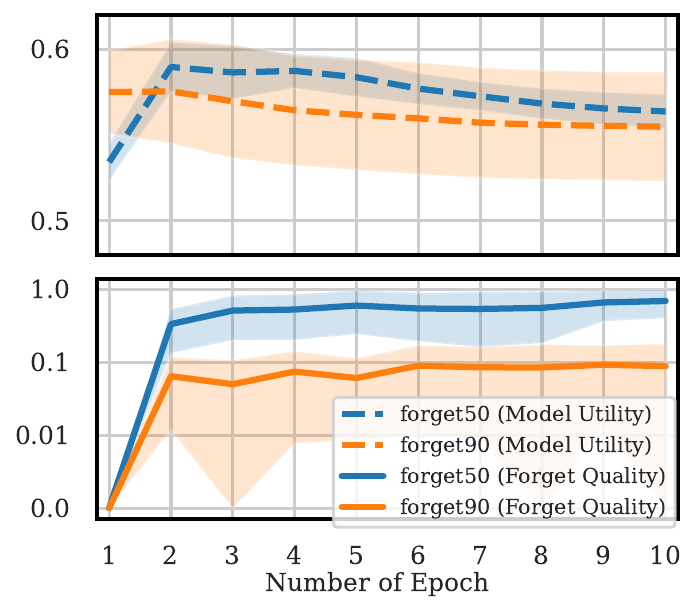}
  \caption{Evolution of forget quality and model utility on Forget50 and Forget90 for NPO+RT with proper componential weights between loss terms. We tune the coefficient of the retain loss term and keep a unit coefficient for the NPO term. For Forget50, we set the coefficient of the retain loss term to be $5.0$ while in Forget90, we set $12.0$.}
  \label{fig.forget50.90}
\end{figure}

\paragraph{Pushing towards the limit: forgetting 50\% - 90\% of TOFU.} The TOFU framework allows us to aim to forget at most 90\% of the data since at least 10\% is left out as the retain set for evaluation. We thus ask the question of whether there exist methods that could effectively forget 50\%-90\% of the TOFU data. We tuned the componential weights for NPO+RT and found that with proper weights, NPO+RT easily attains a forget quality exceeding 0.05 and model utility above 0.55 on Forget50 and Forget90, as reported in Figure~\ref{fig.forget50.90}.

\section{Conclusion}
We propose Negative Preference Optimization (NPO), a simple objective for LLM unlearning. NPO makes steps towards addressing the catastrophic collapse issue in the gradient ascent method. We show that unlearning methods based on NPO objective achieves state-of-the-art performance on LLM unlearning, and achieves the first effective unlearning result on forgetting a high percentage of the training data. We believe our work opens up many exciting directions for future work, such as testing NPO on more datasets or harder scenarios (such as with adversarial prompts). It may also be of interest to generalize the algorithm principle of NPO (preference optimization with negative examples only) to other problems beyond unlearning. 
\section*{Acknowledgement} 
Song Mei is supported by NSF DMS-2210827, CCF-2315725, NSF Career DMS-2339904, ONR N00014-24-S-B001, an Amazon Research Award, and a Google Research Scholar Award. The authors would like to thank Baihe Huang, Xuelin Yang for the valuable discussions. The authors would like to thank Jiantao Jiao for sharing his GPU resources. This research was supported by the Center for AI Safety Compute Cluster. Any opinions, findings, and conclusions or recommendations expressed in this material are those of the authors and do not necessarily reflect the views of the sponsors.

\bibliography{colm2024_conference}
\bibliographystyle{colm2024_conference}

\appendix

\clearpage

\section{Proofs}
\subsection{Proof of Proposition~\ref{prop:npo_conv_to_ga}}\label{pf:prop:npo_conv_to_ga}

Adopt the shorthand $\logratio_i=\log\frac{\policy_\Par(\yforget_i\mid\xforget_i)}{\policy_{\refm}(\yforget_i\mid\xforget_i)}$. 
For any $i\in[\nforget]$, we have
    \begin{align*} 
       \lim_{\itemp\to0} -\frac{2}{\itemp}\cdot\log\sigmoidshort\Bigg(-\itemp\cdot\log\frac{\policy_\Par(\yforget_i\mid\xforget_i)}{\policy_{\refm}(\yforget_i\mid\xforget_i)}\Bigg)-\frac{2}{\beta}\log 2
       &=  \lim_{\itemp\to0} -\frac{2}{\itemp}\cdot\log\sigmoidshort(-\itemp\logratio_i)-\frac{2}{\beta}\log 2\\
       &=\lim_{\itemp\to0} \frac{2}{\itemp}\cdot\log\Big(\frac{1+\exp(\itemp\logratio_i)}{2}\Big)\\
        &=\lim_{\itemp\to0} \frac{2}{\itemp}\cdot\log\Big(1+\frac{\exp(\itemp\logratio_i)-1}{2}\Big)\\
        &=\lim_{\itemp\to0}\frac{2}{\itemp}\cdot\frac{\exp(\itemp\logratio_i)-1}{2}\\
        &=\logratio_i.
    \end{align*}
    Averaging over all $i\in[\nforget]$ and noting that $\sum_{i=1}^\nforget \logratio_i/\nforget =\Loss_{\GA}(\Par)-\E_{\dsetf}[\log\policy_{\refm}(\yforget_i\mid\xforget_i)]$ yields the first part of Propostion~\ref{prop:npo_conv_to_ga}.
    
    For the second part of Propostion~\ref{prop:npo_conv_to_ga}, by definition
    \begin{align*}
        \nabla_\Par\Loss_{\NPO,\itemp}(\Par)
        &=
        \frac{1}{\nforget}\sum_{i=1}^\nforget -\frac{2}{\itemp}\nabla_\Par\log\sigmoidshort(-\itemp\logratio_i)\\
        &=  \frac{1}{\nforget}\sum_{i=1}^\nforget \frac{2}{\itemp}\nabla_\Par\log(1+\exp(\itemp\logratio_i))\\ 
         &=  \frac{1}{\nforget}\sum_{i=1}^\nforget  \frac{2}{\itemp}\cdot\frac{\itemp\exp(\itemp\logratio_i)}{1+\exp(\itemp\logratio_i)}\cdot\nabla_\Par\logratio_i\\ 
          &=  \frac{1}{\nforget}\sum_{i=1}^\nforget \frac{2\exp(\itemp\logratio_i)}{1+\exp(\itemp\logratio_i)}\cdot\nabla_\Par\logratio_i.
    \end{align*}
    Therefore, it follows immediately that
    \begin{align*}
        \lim_{\itemp\to0} \nabla_\Par\Loss_{\NPO,\itemp}(\Par) 
        &=  \lim_{\itemp\to0}\frac{1}{\nforget}\sum_{i=1}^\nforget \frac{2\exp(\itemp\logratio_i)}{1+\exp(\itemp\logratio_i)}\cdot\nabla_\Par\logratio_i= \frac{1}{\nforget}\sum_{i=1}^\nforget \nabla_\Par\logratio_i\\
        &=  \frac{1}{\nforget}\sum_{i=1}^\nforget \nabla_\Par\log\policy_\Par(\yforget_i\mid\xforget_i)=\nabla_\Par\Loss_{\GA}(\Par).
    \end{align*}
This completes the proof of Proposition~\ref{prop:npo_conv_to_ga}.

\subsection{Proof of Theorem~\ref{thm:conv_speed}}\label{sec:pf_thm:conv_speed}
Let $\lrate:=\lrateinit/\nforget$ denote the normalized learning rate. For $i,j\in[\nforget]$, define $\corre_{i,j}:=\<\xforget_i,\xforget_j\>$ and $\pconst_{\init,i}:=\<\Par_\init,\xforget_i\>$. Throughout the proof, we also use $\polyshort_k (k=0,1,2,\ldots)$ to denote constants that may depend on $(\boundinit,\lboundx,\boundx,\itemp)$ but not on $(t,\nforget,\Dim)$. 
By the definition of the logistic model and some algebra, we have 
\begin{align*}
    \nabla_\Par\log\policy_{\Par}(\yforget_i\mid\xforget_i)=\xforget_i(2\yforget_i-1)(1-\policy_\Par(\yforget_i\mid\xforget_i)).
\end{align*} Therefore, the gradient of $\Loss_\GA$ and $\Loss_\NPO$ both lie in the span of $\xforget_1,\ldots,\xforget_{\nforget}$. Consequently,  $\Par^\tth_{\GA}-\Par_\init$ and $\Par^\tth_{\NPO}-\Par_\init$ can be rewritten as follows:
\begin{align*}
    \Par^\tth_{\GA}-\Par_\init=\sum_{i=1}^{\nforget} \direc_{\GA,i}^\tth\cdot\xforget_i,~~~ \Par^\tth_{\NPO}-\Par_\init=\sum_{i=1}^{\nforget} \direc_{\NPO,i}^\tth\cdot\xforget_i,
\end{align*}
where 
\begin{align*}
    \direc_{\GA,i}^\tponeth:= \direc_{\GA,i}^\tth-
    \lrate(2\yforget_i-1)(1-\policy_\Par(\yforget_i\mid\xforget_i)),~~~ \direc_{\NPO,i}^\tponeth:= \direc_{\NPO,i}^\tth-\lrate(2\yforget_i-1)(1-\policy_\Par(\yforget_i\mid\xforget_i))\cdot\NPOweight^\tth_i,
\end{align*}
and 
\begin{align*}\NPOweight^\tth_i=
{\policy_{\Par^\tth_\NPO}^\itemp(\yforget_i\mid\xforget_i)}\big/{[\policy_{\Par^\tth_\NPO}^\itemp(\yforget_i\mid\xforget_i)+\policy_{\Par_\init}^\itemp(\yforget_i\mid\xforget_i)]}.
\end{align*}
We also define $\direc_{\star,i}^{\zeroth}=0$ and 
adopt the shorthand notations $\direc^\tth_\star=(\direc^\tth_{1,\star},\ldots,\direc^\tth_{\nforget,\star})$ for $\star\in\{\GA,\NPO\}$ and $\corre_i=(\corre_{i,1},\ldots,\corre_{i,\nforget})$.  

For $\star\in\{\GA,\NPO\}$, we have 
\begin{align*}
\policy_{\Par^\tth_{\star}}(\yforget_i\mid\xforget_i)
&=\frac{1}{1+\exp\Big((1-2\yforget_i)\<\x_i,\Par^\tth_{\star}\>\Big)}
=\frac{1}{1+\exp\Big((1-2\yforget_i)\pconst_{\init,i}+(1-2\yforget_i)\<\x_i,\sum_{j=1}^\nforget  \direc_{\star}^\tth\xforget_j\>\Big)}\\
&=\frac{1}{1+\exp\Big((1-2\yforget_i)\pconst_{\init,i}+(1-2\yforget_i)\<\direc^\tth_{\star},\corre_i\>\Big)}=:\pred_i\Big(\<\direc^\tth_{\star},\corre_i\>\Big),
\end{align*}
where we denote the dependence on $(\pconst_{\init,i},\yforget_i)$ implicitly using $\pred_i$ for notational simplicity.

Therefore,  letting $\direcb^\tth_{\star,i}:=\<\direc^\tth_{\star},\corre_i\> $ for $\star\in\{\GA,\NPO\}$ and combining all $i\in[\nforget]$, we obtain
 \begin{align*}
      \direcb_{\GA}^\tponeth:= \direcb_{\GA}^\tth-\lrate\cdot\Corre\diff_{\GA}^\tth,~~~ \direcb_{\NPO}^\tponeth:= \direcb_{\NPO}^\tth-\lrate\cdot\Corre\diag\Big\{\diff_{\NPO}^\tth\Big\}\NPOweight^\tth,
 \end{align*}
 where 
 \begin{align*}\Corre&:=(\corre_{i,j})_{1\leq i,j\leq\nforget},
 \\
 \diff_{\star}^\tth&:=\Big(\diff_{\star,1}^\tth,\ldots,\diff_{\star,\nforget}^\tth\Big)^\top,~~
 ~~\diff_{\star,i}^\tth:=(2\yforget_i-1)(1-\pred_i(\direcb_{\star,i}^\tth)),~~~\text{for }i\in[\nforget], \star\in\{\GA,\NPO\},
 \\ 
  \NPOweight^\tth&:=(\NPOweight^\tth_1,\ldots,\NPOweight_\nforget)^\top,
~~~\NPOweight_i^\tth=\NPOweight_i\Big(\direcb^\tth_{\NPO}\Big):={\policy_{\Par^\tth_\NPO}^\itemp(\yforget_i\mid\xforget_i)}\big/{[\policy_{\Par^\tth_\NPO}^\itemp(\yforget_i\mid\xforget_i)+\policy_{\Par_\init}^\itemp(\yforget_i\mid\xforget_i)]}.
\end{align*}
Again, we hide here the dependence on $(\itemp,\pconst_{\init,i},\yforget_i)$ in $\NPOweight_i$ for simplicity.
 We now claim the following results of which the proofs are deferred to Section~\ref{sec:pf_lm:ga_dynamics}~and~\ref{sec:pf_lm:npo_dynamics}.
 \begin{lemma}[$\GA$ converges to infinity linearly ]\label{lm:ga_dynamics}
    Under the assumptions in Theorem~\ref{thm:conv_speed} and the notations in the proof of Theorem~\ref{thm:conv_speed}, there exist some $(\boundinit,\lboundx,\boundx)$-dependent constants $\polyshort_0,\polyshort_1,\polyshort_2>0$   such that the $\GA$ iterations $\{\direcb^\tth_{\GA}\}_{t=1}^\infty$ satisfy 
     \begin{align*}
     \polyshort_1 \lrate t &\leq \direcb_{\GA,i}^\tth\leq  \polyshort_2 \lrate t, \text{ when } y_i=0, \\
     -\polyshort_2 \lrate t &\leq \direcb_{\GA,i}^\tth\leq  -\polyshort_1 \lrate t, \text{ when } y_i=1.
     \end{align*}
     for all $i\in[\nforget]$ and $t\geq 1$ when $\max_{i\neq j}|\corre_{i,j}|\leq \polyshort_0/\nforget.$
 \end{lemma}
  \begin{lemma}[$\NPO$ converges to infinity exponentially slow ]\label{lm:npo_dynamics}
    Under the assumptions in Theorem~\ref{thm:conv_speed} and the notations in the proof of Theorem~\ref{thm:conv_speed}, there exist some $(\boundinit,\lboundx,\boundx,\itemp)$-dependent constants $\polyshort_0,\polyshort_b>0,\polyshort_a\in(0,1)$   such that the $\NPO$ iterations $\{\direcb^\tth_{\NPO}\}_{t=1}^\infty$ satisfy 
     \begin{align*}
     \direcb_{\NPO,i}^\tth&\in[-\frac{1}{\itemp}\log(\polyshort_b\lrate t+1),-\frac{1}{\itemp}\log(\polyshort_a\lrate t+1)] \text{ when } y_i=1,\\
 \direcb_{\NPO,i}^\tth&\in[\frac{1}{\itemp}\log(\polyshort_a\lrate t+1),\frac{1}{\itemp}\log(\polyshort_b\lrate t+1)] \text{ when } y_i=0.
     \end{align*}
     for all $i\in[\nforget]$ and $t\geq 1$ when $\max_{i\neq j}|\corre_{i,j}|\leq \polyshort_0/\nforget.$
 \end{lemma}
 Combining Lemma~\ref{lm:ga_dynamics}~and~\ref{lm:npo_dynamics} and noting
 \begin{align*}
     \|\Par^\tth_\star-\Par_\init\|_{\Xforget^\top\Xforget}&=\sqrt{(\Par^\tth_\star-\Par_\init)^\top {\Xforget^\top\Xforget}(\Par^\tth_\star-\Par_\init)}= \sqrt{\direc^{\tth\top}_\star\Xforget {\Xforget^\top\Xforget}\Xforget^\top\direc^\tth_\star}\\
     &=\sqrt{\direcb^{\tth\top}_\star{(\Xforget^\top\Xforget)}^{-1}\Xforget {\Xforget^\top\Xforget}\Xforget^\top{(\Xforget^\top\Xforget)}^{-1}\direcb^\tth_\star}=\|\direcb^{\tth}_\star\|_2
 \end{align*} for $\star\in\{\GA,\NPO\}$ completes the proof. 

\subsection{Proof of Lemma~\ref{lm:ga_dynamics}}\label{sec:pf_lm:ga_dynamics}
We prove Lemma~\ref{lm:ga_dynamics} by induction. 
\paragraph{Case 1: $t=1$}
When $t=1$, since $|\const_{\init}|\leq \|\xforget_i\|_2\cdot\|\Par_\init\|_2\leq\boundx\boundinit$, it follows from the definition of $\pred_i(\cdot)$ that 
\begin{align*}
\polyshort_3\leq \diff^\zeroth_{\GA,i}\leq 1 \text{ when } y_i=1,~~~-1\leq \diff^\zeroth_{\GA,i}\leq -\polyshort_4  \text{ when } y_i=0
\end{align*} for all $i
\in[\nforget]$ 
for  some constants $\polyshort_3,\polyshort_4\in(0,1)$ depending only on $(\boundinit,\lboundx,\boundx)$. 
Note that
there exists a constant $\polyshort_0>0$ depending on $\polyshort_3,\polyshort_4,\lboundx$ such that
\begin{align*}
\Big|\sum_{j\neq i }\corre_{i,j}\diff_{\GA,j}^\zeroth\Big| \leq \frac{\corre_{i,i}}{2}\Big|\diff_{\GA,i}^\zeroth\Big|
\end{align*} for all $i\in[\nforget]$ when $\max_{i\neq j }|\corre_{i,j}|\leq \polyshort_0/\nforget$. It follows that 
\begin{align*}
 -\lrate\cdot\corre_i^\top\diff_{\GA}^\zeroth&\in[-\frac32\corre_{i,i}\lrate|\diff_{\GA,i}^\zeroth|,-\frac12 \corre_{i,i}
     \lrate|\diff_{\GA,i}^\zeroth|]\in[-\polyshort_2\lrate,-\polyshort_1\lrate]~ \text{ when } y_i=1,\\   
      -\lrate\cdot\corre_i^\top\diff_{\GA}^\zeroth&\in[\frac12\corre_{i,i}\lrate|\diff_{\GA,i}^\zeroth|,\frac32 \corre_{i,i}
     |\lrate\diff_{\GA,i}^\zeroth|]\in[\polyshort_1\lrate,\polyshort_2\lrate]~ \text{ when } y_i=0
\end{align*} 
for some $(\boundinit,\lboundx,\boundx)$-dependent constants $\polyshort_1,\polyshort_2>0.$
Therefore, \begin{align*}    \direcb_{\GA,i}^\first&
= \direcb_{\GA,i}^\zeroth-\lrate\cdot\corre_i^\top\diff_{\GA}^\zeroth \in[-\polyshort_2\lrate,-\polyshort_1\lrate]~ \text{ when } y_i=1,\\
 \direcb_{\GA,i}^\first&
= \direcb_{\GA,i}^\zeroth-\lrate\cdot\corre_i^\top\diff_{\GA}^\zeroth \in[\polyshort_1\lrate,\polyshort_2\lrate]~ \text{ when } y_i=0.
\end{align*} As a consequence,
\begin{align*}
      \direcb_{\GA,i}^\first&\leq \direcb_{\GA,i}^\zeroth=0,~~ \diff_{\GA,i}^\first\geq \diff_{\GA,i}^\zeroth~ \text{ when } y_i=1,\\
\direcb_{\GA,i}^\first&\geq \direcb_{\GA,i}^\zeroth=0,~~ \diff_{\GA,i}^\first\leq \diff_{\GA,i}^\zeroth~ \text{ when } y_i=0.
\end{align*}
\paragraph{Case 2: $t=K+1$}
Now, suppose we have 
\begin{align*}    \direcb_{\GA,i}^\tth&
 \in[-\polyshort_2\lrate t,-\polyshort_1\lrate t]~ \text{ when } y_i=1,\\
 \direcb_{\GA,i}^\tth&
\in[\polyshort_1\lrate t,\polyshort_2\lrate t]~ \text{ when } y_i=0
\end{align*} for $t\in[K]$
and 
\begin{align*}
      \direcb_{\GA,i}^\Kth&\leq\ldots\leq \direcb_{\GA,i}^\zeroth=0,~~ \diff_{\GA,i}^\Kth\geq\ldots\geq \diff_{\GA,i}^\zeroth~ \text{ when } y_i=1,\\
\direcb_{\GA,i}^\Kth&\geq\ldots\geq \direcb_{\GA,i}^\zeroth=0,~~ \diff_{\GA,i}^\Kth\leq\ldots\leq \diff_{\GA,i}^\zeroth~ \text{ when } y_i=0.
\end{align*}
By the monotonicity of $\diff^\tth_{\GA,i}$, we have 
\begin{align*}
\polyshort_3\leq \diff^\Kth_{\GA,i}\leq 1 \text{ when } y_i=1,~~~-1\leq \diff^\Kth_{\GA,i}\leq -\polyshort_4  \text{ when } y_i=0
\end{align*} for all $i
\in[\nforget]$. Therefore, following similar arguments as in the $t=1$ case, we have 
\begin{align*}
 -\lrate\cdot\corre_i^\top\diff_{\GA}^\Kth&\in[-\polyshort_2\lrate,-\polyshort_1\lrate]~ \text{ when } y_i=1,\\ 
      -\lrate\cdot\corre_i^\top\diff_{\GA}^\Kth&\in[\polyshort_1\lrate,\polyshort_2\lrate]~ \text{ when } y_i=0.
\end{align*} 
Then it follows from the induction assumption that
\begin{align*}    \direcb_{\GA,i}^\Kponeth&
= \direcb_{\GA,i}^\Kth-\lrate\cdot\corre_i^\top\diff_{\GA}^\Kth \in[-\polyshort_2\lrate(K+1),-\polyshort_1\lrate(K+1)]~ \text{ when } y_i=1,\\
 \direcb_{\GA,i}^\Kponeth&
= \direcb_{\GA,i}^\Kth-\lrate\cdot\corre_i^\top\diff_{\GA}^\Kth \in[\polyshort_1\lrate(K+1),\polyshort_2\lrate(K+1)]~ \text{ when } y_i=0, 
\end{align*} 
    and also
    \begin{align*}
         \direcb_{\GA,i}^\Kponeth&\leq \direcb_{\GA,i}^\Kth,~~ \diff_{\GA,i}^\Kponeth\geq \diff_{\GA,i}^\Kth~ \text{ when } y_i=1,\\
\direcb_{\GA,i}^\Kponeth&\geq \direcb_{\GA,i}^\Kth,~~ \diff_{\GA,i}^\Kponeth\leq \diff_{\GA,i}^\Kth~ \text{ when } y_i=0.
    \end{align*} This concludes the induction step and therefore completes the proof.

\subsection{Proof of Lemma~\ref{lm:npo_dynamics}}\label{sec:pf_lm:npo_dynamics}
    We prove Lemma~\ref{lm:npo_dynamics} by induction. 
    
    Our induction assumption is the following: there exist some constants $\polyshort_0>0,\polyshort_a\in(0,1),\polyshort_b>0,\polyshort_1,\polyshort_2$ depending only on $(\boundinit,\lboundx,\boundx,\itemp)$ such that when $\max_{i\neq j}|\corre_{i,j}|\leq\polyshort_0/\nforget$, for any $t\geq 1$ we have
    \begin{enumerate}
        \item \begin{align*}
      \direcb_{\NPO,i}^\tth&\leq\ldots\leq \direcb_{\NPO,i}^\zeroth=0,~~ \diff_{\NPO,i}^\tth\geq\ldots\geq \diff_{\NPO,i}^\zeroth~ \text{ when } y_i=1,\\
\direcb_{\NPO,i}^\tth&\geq\ldots\geq \direcb_{\NPO,i}^\zeroth=0,~~ \diff_{\NPO,i}^\tth\leq\ldots\leq \diff_{\NPO,i}^\zeroth~ \text{ when } y_i=0.
\end{align*}
  \item \begin{align}
      \direcb_{\NPO,i}^\tth&\in[-\frac{1}{\itemp}\log(\polyshort_b\lrate t+1),-\frac{1}{\itemp}\log(\polyshort_a\lrate t+1)] \text{ when } y_i=1,\label{eq:npo_induction_-1}\\
 \direcb_{\NPO,i}^\tth&\in[\frac{1}{\itemp}\log(\polyshort_a\lrate t+1),\frac{1}{\itemp}\log(\polyshort_b\lrate t+1)] \text{ when } y_i=0.\label{eq:npo_induction_0}
\end{align}
  \item \begin{align}
      \direcb_{\NPO,i}^\tth&\in[-\polyshort_2-\frac{1}{\itemp}\log (\lrate t+1),-\polyshort_1-\frac{1}{\itemp}\log (\lrate t+1)] \text{ when } y_i=1,\label{eq:npo_induction_1}\\
\direcb_{\NPO,i}^\tth&\in[\polyshort_1+\frac{1}{\itemp}\log (\lrate t+1),\polyshort_2+\frac{1}{\itemp}\log (\lrate t+1)] \text{ when } y_i=0.\label{eq:npo_induction_2}
\end{align}

    \end{enumerate}
    Lemma~\ref{lm:npo_dynamics} follows immediately from the second part of the induction assumption. 
    
In the following, we first specify the parameter-dependent constants $\polyshort_0,\polyshort_1,\polyshort_2,\polyshort_a,\polyshort_b$ in the $t=1$ case and prove the induction assumption when $k=1$. Then given the induction assumption holds when $t\leq K$, we prove that it holds when $t=K+1$ as well.
\paragraph{Case 1: $t=1$}
When $t=1$, since $|\const_{\init}|\leq \|\xforget_i\|_2\cdot\|\Par_\init\|_2\leq\boundx\boundinit$, it follows from the definition of $\pred_i(\cdot)$ that 
\begin{align}
\polyshort_3\leq \diff^\zeroth_{\NPO,i}\leq 1 \text{ when } y_i=1,~~~-1\leq \diff^\zeroth_{\NPO,i}\leq -\polyshort_4  \text{ when } y_i=0\label{eq:npo_delta_bound1}
\end{align} for all $i
\in[\nforget]$ 
for  some constants $\polyshort_3,\polyshort_4\in(0,1)$ depending only on $(\boundinit,\lboundx,\boundx)$. Moreover, we claim that
\begin{align}
     \NPOweight^\tth_i\in\Big[\polyshort_5\cdot \exp\Big((2\yforget_i-1)\itemp\direcb^\tth_{\NPO,i}\Big),\polyshort_6\cdot \exp\Big((2\yforget_i-1)\itemp\direcb^\tth_{\NPO,i}\Big)
     \label{claim:npoweight_1}
   \Big] 
\end{align}
for all $i\in[\nforget]$ and $t$ such that $\pred_i(\direcb^\tth_{\NPO,i})\leq\pred_i(\direcb^\zeroth_{\NPO,i})$ for some $(\boundinit,\lboundx,\boundx,\itemp)$-dependent constants $\polyshort_5,\polyshort_6>0$. 

Now, suppose Eq.~(\ref{eq:npo_induction_1})~and~~(\ref{eq:npo_induction_2}) hold for some  $(\boundinit,\lboundx,\boundx,\itemp)$-dependent constants $\polyshort_1,\polyshort_2>0$ which we will specify later. 
Then, 
there exists a constant $\polyshort_0>0$ depending on $\polyshort_{1:6},\lboundx$ such that
\begin{align}
\Big|\sum_{j\neq i }\corre_{i,j}\diff_{\NPO,j}^\zeroth\NPOweight_j\Big| \leq \frac{\corre_{i,i}}{2}\Big|\diff_{\NPO,i}^\zeroth\NPOweight_i\Big|\label{eq:npo_off_diag_1}
\end{align} for all $i\in[\nforget]$ when $\max_{i\neq j }|\corre_{i,j}|\leq \polyshort_0/\nforget$. Furthermore, combining Eq.~(\ref{eq:npo_delta_bound1}),~(\ref{claim:npoweight_1}),~(\ref{eq:npo_off_diag_1}) gives
\begin{align*}
 -\lrate\cdot\corre_i^\top\diag\{\diff_{\NPO}^\zeroth\}\NPOweight_i^\zeroth
 &\in\Big[-\frac32\corre_{i,i}\lrate|\diff_{\NPO,i}^\zeroth|\NPOweight_i^\zeroth,-\frac12 \corre_{i,i}
     \lrate|\diff_{\NPO,i}^\zeroth|\NPOweight_i^\zeroth\Big]\\
     &\in[-\polyshort_8\lrate\exp(\itemp\direcb^\zeroth_{\NPO,i}),-\polyshort_7\lrate\exp(\itemp\direcb^\zeroth_{\NPO,i})]~ \text{ when } y_i=1,\\   
  -\lrate\cdot\corre_i^\top\diag\{\diff_{\NPO}^\zeroth\}\NPOweight_i^\zeroth
 &\in\Big[\frac12\corre_{i,i}\lrate|\diff_{\NPO,i}^\zeroth|\NPOweight_i^\zeroth,\frac32 \corre_{i,i}
     \lrate|\diff_{\NPO,i}^\zeroth|\NPOweight_i^\zeroth\Big]\\
     &\in[\polyshort_7\lrate\exp(-\itemp\direcb^\zeroth_{\NPO,i}),\polyshort_8\lrate\exp(\itemp\direcb^\zeroth_{\NPO,i})]~ \text{ when } y_i=0,
\end{align*} 
if $\max_{i\neq j }|\corre_{i,j}|\leq \polyshort_0/\nforget.$
Here  $\polyshort_7,\polyshort_8>0$ are some $(\polyshort_{3},\polyshort_{4},\polyshort_{5},\polyshort_{6},\itemp)$-dependent constants, and we pick $\polyshort_7$ such that $\polyshort_7\itemp<1$.  As a consequence,
\begin{align*}
      \direcb_{\NPO,i}^\first&\leq \direcb_{\NPO,i}^\zeroth=0,~~ \diff_{\NPO,i}^\first\geq \diff_{\NPO,i}^\zeroth~ \text{ when } y_i=1,\\
\direcb_{\NPO,i}^\first&\geq \direcb_{\NPO,i}^\zeroth=0,~~ \diff_{\NPO,i}^\first\leq \diff_{\NPO,i}^\zeroth~ \text{ when } y_i=0.
\end{align*} This concludes the proof of the first part of the induction assumption.

Now, we start to prove the second part of the induction assumption. 
For $i$ such that $\yforget_i=1$,
consider the ordinary differential equations
\begin{align*}
    \direcb_l'(t) &= -\polyshort_8\lrate(1+\exp(\polyshort_8))\cdot\exp(\itemp\direcb_l(t)),~~~~~~ \direcb_l(0) = 0;\\
     \direcb_u'(t) &= -\polyshort_7\lrate\cdot\exp(\itemp\direcb_l(t)), ~~~~~~\direcb_u(0) = 0.
\end{align*}
It can be verified that the ODEs have closed-form solutions
\begin{align*}
    \direcb_l(t)&=-\frac{1}{\itemp}\log(\itemp\polyshort_8\lrate(1+\exp(\polyshort_8))t+1),\\
    \direcb_u(t)&=-\frac{1}{\itemp}\log(\itemp\polyshort_7\lrate t+1).
\end{align*}
Since
\begin{align*}\direcb_{\NPO,i}^\first= \direcb_{\NPO,i}^\first-\lrate\cdot\corre_i^\top\diff_{\NPO,i}^\zeroth\geq \direcb_{\NPO,i}^\first -\polyshort_8\lrate\exp(\itemp\direcb^\zeroth_{\NPO,i})\geq \direcb_{\NPO,i}^\first -\polyshort_8\lrate,\end{align*} it follows that for any point $\direcb^{\epsth}_{\NPO,i}=\eps\direcb^{\first}_{\NPO,i}+(1-\eps)\direcb^{\zeroth}_{\NPO,i}$ with $\eps\in[0,1]$ 
\begin{align*}
-\polyshort_8\lrate(1+\exp(\polyshort_8\lrate))\cdot\exp(\itemp\direcb^{\epsth}_{\NPO,i})
&\leq -\polyshort_8\lrate\exp(\itemp\direcb^\zeroth_{\NPO,i})
\leq
-\polyshort_7\lrate\exp(\itemp\direcb^\zeroth_{\NPO,i})\\
&\leq
-\polyshort_7\lrate\cdot\exp(\itemp\direcb^{\epsth}_{\NPO,i}).
\end{align*}
Therefore, we have by the comparison theorem for ODEs that
\begin{align*}
    \direcb_l(\eps) \leq \direcb_{\NPO,i}^\epsth\leq    \direcb_u(\eps)
\end{align*} for $\eps\in[0,1]$. Setting \begin{align*}\polyshort_a := \polyshort_7\itemp,~~~~~\polyshort_b := \itemp\polyshort_8(1+\exp(\polyshort_8))\end{align*} concludes the second part of the induction assumption.

For the last part of the induction assumption, 
since $\log(x+1)+\log(c)\leq \log(cx+1)\leq \log(x+1)+\log(c+1)$ when $c\leq 1$, we have 
\begin{align*}
    -\frac{1}{\itemp}[\log(\lrate t+1)+\log(\itemp\polyshort_8(1+\exp(\polyshort_8)))] \leq \direcb_l(t)\leq\direcb_u(t)\leq  -\frac{1}{\itemp}[\log(\lrate t+1)+\log(\itemp\polyshort_7)], 
\end{align*}
where the last inequality uses $\itemp\polyshort_7<1$. Therefore, we obtain
\begin{align*}
    \direcb^\first_{\NPO,i}\in[ -\frac{1}{\itemp}\log(\lrate t+1)-\polyshort_2,-\frac{1}{\itemp}\log(\lrate t+1)-\polyshort_1],
\end{align*}
where
\begin{align*}
\polyshort_1:=\frac{1}{\itemp}\log(\itemp\polyshort_7),~~~~~
    \polyshort_2:= \frac{1}{\itemp}\log(\itemp\polyshort_8(1+\exp(\polyshort_8)))
\end{align*} for $i$ such that $\yforget_i=1$. Following the same arguments, similarly, we also have
\begin{align*}
    \direcb^\first_{\NPO,i}\in[ \frac{1}{\itemp}\log(\lrate t+1)+\polyshort_1,\frac{1}{\itemp}\log(\lrate t+1)+\polyshort_2]
\end{align*} for $i$ such that $\yforget_i=0$. This concludes the last part of the induction assumption.
\paragraph{Case 2: $t=K+1$}
Suppose the induction assumption holds for $t\in[K]$, we now show that the induction assumption holds for $t=K+1$ as well. Following the proof of $t=1$ case and using the monotonicity property of $\{\diff_{\NPO,i}^\tth\}_{t=0}^K$, we have
\begin{align*}
\polyshort_3\leq \diff^\Kth_{\NPO,i}\leq 1 \text{ when } y_i=1,~~~-1\leq \diff^\Kth_{\NPO,i}\leq -\polyshort_4  \text{ when } y_i=0
\end{align*} for all $i
\in[\nforget]$. Since $\pred_i(\direcb^\Kth_{\NPO,i})\leq\pred_i(\direcb^\Kth_{\NPO,i})$ by the definition of $\pred_i$ and the monotonicity of $\{\direcb^\tth_{\NPO,i}\}_{t=0}^K$, it follows from  Claim~(\ref{claim:npoweight_1}) that
\begin{align*}
     \NPOweight^\Kth_i\in\Big[\polyshort_5\cdot \exp\Big((2\yforget_i-1)\itemp\direcb^\Kth_{\NPO,i}\Big),\polyshort_6\cdot \exp\Big((2\yforget_i-1)\itemp\direcb^\Kth_{\NPO,i}\Big)
   \Big]. 
\end{align*}
Putting the last two displays together and following the same argument as the $t=1$ case, we find that
\begin{align*}
    -\lrate\cdot\corre_i^\top\diag\{\diff_{\NPO}^\Kth\}\NPOweight_i^\Kth
     &\in[-\polyshort_8\lrate\exp(\itemp\direcb^\Kth_{\NPO,i}),-\polyshort_7\lrate\exp(\itemp\direcb^\Kth_{\NPO,i})]~ \text{ when } y_i=1,\\   
  -\lrate\cdot\corre_i^\top\diag\{\diff_{\NPO}^\Kth\}\NPOweight_i^\Kth
   &\in[\polyshort_7\lrate\exp(-\itemp\direcb^\Kth_{\NPO,i}),\polyshort_8\lrate\exp(\itemp\direcb^\Kth_{\NPO,i})]~ \text{ when } y_i=0.
\end{align*} The first part of the induction assumption (for $t=K+1$) follows immediately as the sign of the gradient updates $-\lrate\cdot\corre_i^\top\diag\{\diff_{\NPO}^\Kth\}\NPOweight_i^\Kth$ are determined as above.

Note that $\direcb_l(K) \leq \direcb_{\NPO,i}^{\Kth}\leq   \direcb_u(K)$ by the induction assumption. 
Similarly, using the comparison theorem of ODEs, we obtain
\begin{align*}
    \direcb_l(K+\eps) \leq \direcb_{\NPO,i}^{(K+\eps)}\leq    \direcb_u(K+\eps)
\end{align*} for $\eps\in[0,1]$ and $\direcb_{\NPO,i}^{(K+\eps)}:= \eps \direcb_{\NPO,i}^{\Kponeth}+(1-\eps)\direcb_{\NPO,i}^{\Kth}$. Choosing $\eps=1$ gives the second part of the induction assumption for $t=K+1$. The last part of the induction assumption for $t=K+1$ follows from the same algebra as in the $t=1$ case.

    \paragraph{Proof of Claim~(\ref{claim:npoweight_1})}
By definition
\begin{align*}
    \NPOweight^\tth_i=\frac{\policy_{\Par^\tth_\NPO}^\itemp(\yforget_i\mid\xforget_i)}{\policy_{\Par^\tth_\NPO}^\itemp(\yforget_i\mid\xforget_i)+\policy_\refm^\itemp(\yforget_i\mid\xforget_i)}=\frac{\pred_i(\direcb^\tth_{\NPO,i})^\itemp}{\pred_i(\direcb^\tth_{\NPO,i})^\itemp+\pred_i(\direcb^\zeroth_{\NPO,i})^\itemp}.
\end{align*}
When $\pred_i(\direcb^\tth_{\NPO,i})\leq\pred_i(\direcb^\zeroth_{\NPO,i})$, we have
\begin{align}
\NPOweight^\tth_i\in\Bigg[\frac{\pred_i(\direcb^\tth_{\NPO,i})^\itemp}{2\pred_i(\direcb^\zeroth_{\NPO,i})^\itemp},\frac{\pred_i(\direcb^\tth_{\NPO,i})^\itemp}{\pred_i(\direcb^\zeroth_{\NPO,i})^\itemp}\Bigg]\in[\polyshort_l\cdot\pred_i(\direcb^\tth_{\NPO,i})^\itemp,\polyshort_u\cdot\pred_i(\direcb^\tth_{\NPO,i})^\itemp]\label{eq:npoweight_c_1}
\end{align}
for some constants $\polyshort_l,\polyshort_u>0$ depending only on $(\boundinit,\lboundx,\boundx,\itemp)$. Note that $\pred_i(\direcb^\tth_{\NPO,i})\leq\pred_i(\direcb^\zeroth_{\NPO,i})$ is equivalent to $(1-2\yforget_i)\direcb^\tth_{\NPO,i}\geq 0$. Therefore, under this condition, we have
\begin{align}
    \pred_i(\direcb^\tth_{\NPO,i})
    &= 
    \frac{1}{1+\exp\Big((1-2\yforget_i)\pconst_{\init,i}+(1-2\yforget_i)\direcb^\tth_{\NPO,i}\Big)}\notag \\
   & =
     \frac{\exp\Big((2\yforget_i-1)\direcb^\tth_{\NPO,i}\Big)}{\exp\Big((2\yforget_i-1)\direcb^\tth_{\NPO,i}\Big)+\exp\Big((1-2\yforget_i)\pconst_{\init,i}\Big)}
   \notag  \\
     &\in\Bigg[
       \frac{\exp\Big((2\yforget_i-1)\direcb^\tth_{\NPO,i}\Big)}{1+\exp\Big((1-2\yforget_i)\pconst_{\init,i}\Big)},  \frac{\exp\Big((2\yforget_i-1)\direcb^\tth_{\NPO,i}\Big)}{\exp\Big((1-2\yforget_i)\pconst_{\init,i}\Big)}
     \Bigg]\label{eq:npoweight_c_2}.
\end{align}
Putting Eq.~(\ref{eq:npoweight_c_1})~and~(\ref{eq:npoweight_c_2}) together and recalling that $
|\pconst_{\init,i}|\leq\boundinit\boundx$, we obtain
\begin{align*}
   \NPOweight^\tth_i\in\Big[\polyshort_5\cdot \exp\Big((2\yforget_i-1)\itemp\direcb^\tth_{\NPO,i}\Big),\polyshort_6\cdot \exp\Big((2\yforget_i-1)\itemp\direcb^\tth_{\NPO,i}\Big)
   \Big] 
\end{align*}
for some constants $\polyshort_5,\polyshort_6>0$ depending only on $(\boundinit,\lboundx,\boundx,\itemp)$. This concludes the proof of Claim~(\ref{claim:npoweight_1}).

\newpage
\section{The role of KL Divergence on the Forget Set}
In this section, we report \emph{Forget KL}, the KL divergence between the output distributions of the initial model and the unlearned model on the forget set, defined as \mbox{$\E_{\dsetf}\mathrm{KL}(\policy_{{\refm}}(\cdot\mid\x)||\policy_\Par(\cdot\mid\x))$}. In experiments on the synthetic dataset and TOFU dataset,  we observe that the models exhibit better unlearning performance when the forget KL is maintained at a moderate level. This suggests that explicitly maximizing the forget KL may not be an ideal objective for unlearning tasks. 

\subsection{Synthetic Experiment}
\begin{figure}[H]
\centering
\includegraphics[width = 0.6\linewidth]{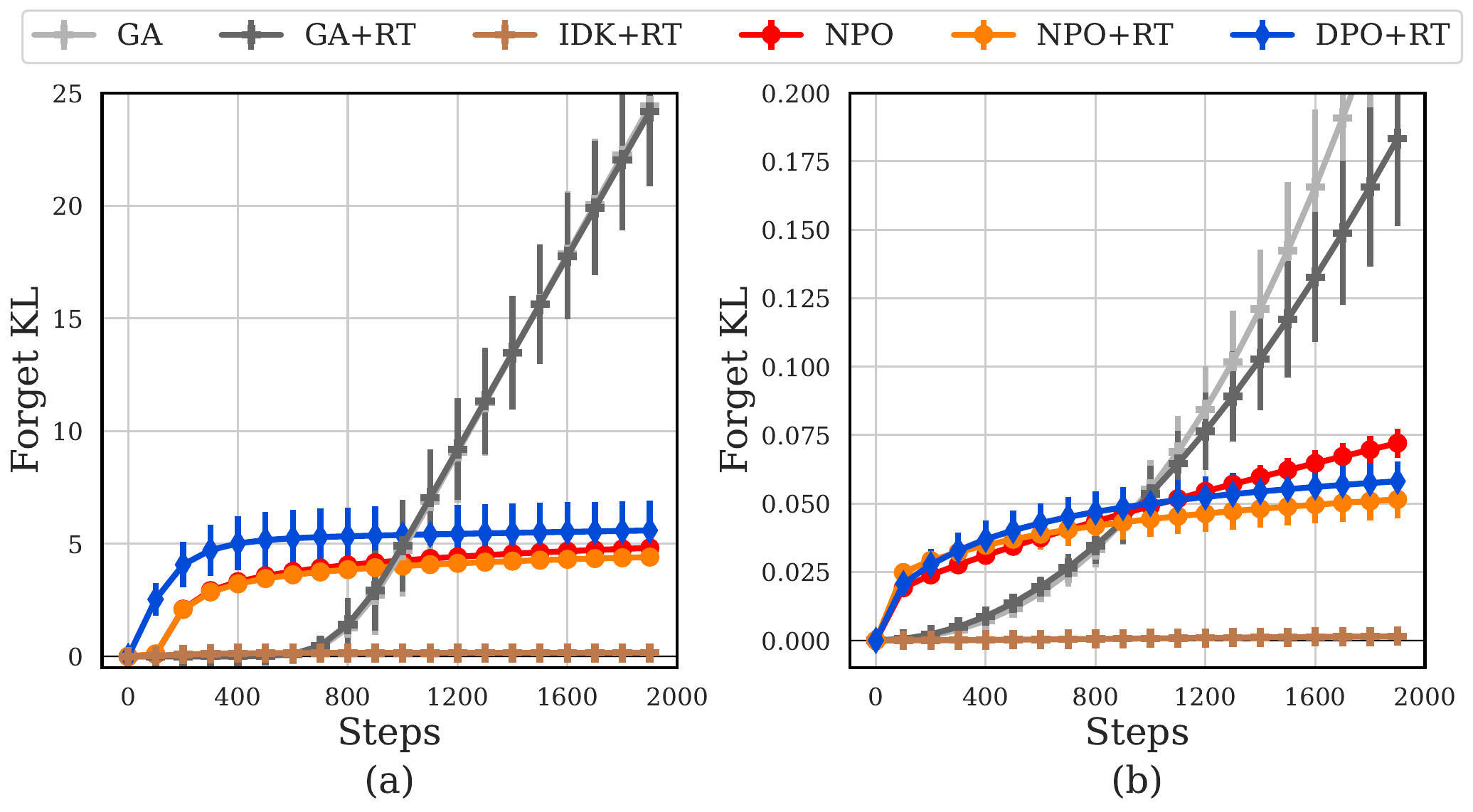}
\caption{Forget KL versus optimization steps for all methods in the synthetic experiment. (a): $\alpha=1$, (b): $\alpha=0$.  The errorbars denote $\pm1$ standard deviation over $5$ runs.}\label{fig:synthetic_2} 
\end{figure}

We present the forget KL for the synthetic experiments in Figure~\ref{fig:synthetic_2}~(a)~and~(b).  Combining with Figure~\ref{fig:synthetic_1}, we find that the unlearned models attain  Pareto frontiers when the forget KL is suitably large---an excessively large forget KL (as in $\GA,\GA+\RT$ after $1200$ steps) or an excessively small forget KL (as in $\IDK+\RT$) may deteriorate the unlearning performance.

\section{Experimental details of the synthetic experiments}
In this section, we discuss the experimental details of the synthetic experiments studied in Section~\ref{sec:synthetic_exp}.
\paragraph{Initial model and retrained model.}
We create the initial model $\policy_{\refm}$ and the retrained model $\policy_{\retrain}$ via optimizing over $\Par$ using the cross-entropy loss on the entire dataset $\cD = \dsetf \cup \dsetr$ and the retain dataset $\dsetr$, respectively. Concretely, initializing at $\Par=\zeros_{128}$, we run gradient descent for $20000$ steps with the learning rate equals $0.05$ to obtain the initial model $\policy_{\refm}$ and the retrained model $\policy_{\retrain}$.

\paragraph{Unlearning.}
During unlearning, starting from the initial model $\policy_{\refm}$, we run gradient descent on each of the loss functions for $2000$ steps with the learning rates selected via a grid search. We choose the learning rates so that the training remains stable within $2000$ steps. It should be noted that variations of the learning rates may affect the number of steps needed to reach the minimal forget distance (or retain distance) in Figure~\ref{fig:synthetic_1}~(a1, a2, b). However, they are less likely to alter the Pareto curves shown in Figure~\ref{fig:synthetic_1}~(c). A grid search is also conducted to select the optimal $\itemp$ for NPO (and DPO)-based methods. The choices of learning rate and $\itemp$ are summarized in Table~\ref{table:lr_synthetic}. 
{
\renewcommand{\arrayrulewidth}{1pt}
\begin{table}[ht]
\centering
\begin{tabular}{lcccc}
\toprule
\multicolumn{1}{c}{Method} & \multicolumn{2}{c}{learning rate}  & \multicolumn{2}{c}{$\itemp$} \\
\cmidrule{2-3}\cmidrule{4-5}
 & $\alpha=1$ & $\alpha=0$  & $\alpha=1$ & $\alpha=0$ \\
\midrule
GA,GA+RT, IDK+RT & 5e-4 & 1e-4 &N/A&N/A\\
\midrule
NPO, NPO+RT & 5e-3 & 5e-2 & 1 &10\\
\midrule
DPO+RT & 5e-3 & 5e-2 & 0.1 &5\\
\bottomrule
\end{tabular}
\caption{Values of learning rate and $\itemp$ for different methods when $\alpha=1$ and $\alpha=0$ in the synthetic experiments.}
\label{table:lr_synthetic}
\end{table}
}

\section{Experiments on the TOFU dataset}
In this section, we provide details of the experiments on the TOFU dataset \citep{maini2024tofu}. We first present a detailed explanation of the metrics, the baseline methods and the hyperparameters in the experiments. Then, we provide the full results on different levels of the tasks.

\subsection{Experiments Setup}\label{appendix.tofu.setup}
\subsubsection{Dataset}\label{appendix.tofu.dataset}
\paragraph{TOFU Dataset.} We evaluate NPO-based methods and all baselines on the TOFU (Task of Fictitious Unlearning) dataset \citep{maini2024tofu} designed for measuring the unlearning methods for LLMs. TOFU contains 200 fictitious author profiles, each consisting of 20 question-answering pairs generated by GPT-4 based on a set of predefined attributes. These profiles are fictitious and do not exist in the pre-training data, providing a controlled environment for studying unlearning LLMs. TOFU introduces three levels of unlearning tasks, each aiming at forgetting a subset of 2, 10, and 20 authors (comprising 1\%, 5\%, and 10\% of the training data, respectively), referred to as the \emph{forget set} $\dsetf$, with a computational constraint that scales linearly with the size of the forget set. We refer to these tasks as Forget01, Forget05, and Forget10, respectively. 

\paragraph{Dataset for Evaluation.}
In addition to the forget set, \citet{maini2024tofu} also introduced other datasets to measure the performance of the unlearned model. The \emph{retain set} $\dsetr$ is the part of the data that we do not hope the model to forget, which is, by definition, the complementary set of the forget set in the full dataset. To evaluate the model performance on the retain set, TOFU earmarks a subset of 400 question-answer pairs, accounting for 10\% of the data, as an exclusive retain set that is never included in the forget set for any task. Moreover, to measure the general capacities of the unlearned models, two additional datasets are introduced: the Real Authors set and the Real World set. The Real Authors set includes question-answer pairs about authors in the real world and often deals with neighboring concepts entangled with those in the forget set. The Real World set contains commonsense knowledge about the real world and is designed to examine the general world knowledge of the unlearned model.

\paragraph{Dataset beyond Forget10.} In scenarios where the targeted forget set exceeds 10\% of the data (Forget20, Forget30, Forget50, and Forget90), we reorganize the original forget and retain sets within the TOFU dataset. To assess the Truth Ratio on an evaluation dataset, it is necessary to utilize the perturbed and paraphrased answers, which in TOFU were generated by properly prompting GPT4. To avoid any potential distribution shift from the newly crafted responses and their original forms, we evaluate the Truth Ratio using the publicly available data within TOFU. Consequently, even for tasks that are beyond forget10, we continue to use the data from the standard forget10 subset to compute the Truth Ratio on the forget set. This serves as a reasonable proxy for evaluating the Truth Ratio on the full forget set.

\subsubsection{Metrics}\label{appendix.tofu.metric}
\paragraph{Model Utility.}
We measure the general capacities of the unlearned model using \emph{Model Utility}, which aggregates multiple metrics across the retain set, the real-world set and the real-author set. Given a question-answer pair $x = [q,a],$ we compute the normalized conditional probability $\P(a \mid q)^{1/|a|},$ where $|\cdot|$ denotes the number of tokens in a certain sequence. This probability is then averaged over the retain set, the Real Authors set, and the Real World set each. We also compute the averaged ROUGE-L recall score \citep{lin2004rouge} across these datasets, a metric that evaluates the accuracy of the model's response compared to the reference answers. Finally, we compute the averaged Truth Ratio on the three datasets above. The Truth Ratio defined in \cite{maini2024tofu} measures how likely the unlearned model will give a correct answer versus an incorrect one. More specifically, given a question $q,$ \citet{maini2024tofu} generated a paraphrased (correct) answer $\widetilde a$ via prompting GPT4. They then generated five perturbed answers with the exactly same templates but incorrect answers $\widehat{a}_i, i = 1,2,3,4,5$ in the same way. The Truth Ratio is defined by
\begin{equation}\label{eqn.def.truth.ratio}
    R_{truth} := \frac{\frac{1}{5}\sum_{i=1}^5 \P(\widehat a_i \mid q)^{1/|\widehat{a}_i|}}{\P(\widetilde{a} \mid q)^{1/|\widetilde{a}|}}.
\end{equation}
The model utility is defined as the harmonic average of the nine metrics above (the probability, the ROUGE score, and the Truth Ratio on the retain set, the Real Authors set, and the Real World set).

\paragraph{Forget Quality.}
Forget quality assesses how well the unlearned model mimics the retrained model (defined as the model trained only on the retain set). 
This is a rigorous measurement as the ultimate goal for unlearning LLM is not only to stop generating the content related to the forget set but also to make the unlearned model indistinguishable from the retrained one. From a practical view of point, this requires the next-token probability given a prefix of the unlearned model to be as close as possible to that of the retrained model. In TOFU, they compute the Truth Ratio (defined in Eq.~\ref{eqn.def.truth.ratio}) on each question-answer pair from the forget set. Instead of simply averaging them, they test whether the distribution of the Truth Ratio computed from the unlearned and the retrained models are indistinguishable. More specifically, they perform the Kolmogorov-Smirnov (KS) test and compute the p-value of the test. A large p-value indicates that the two models are indistinguishable from the Truth Ratio. When the p-value is above 0.05, we say the forgetting is significant.

\subsubsection{Baseline Methods}
In this section, we introduce the baseline methods in our experiments.

\paragraph{GA-based Methods.}
Gradient Ascent (GA) is a key component in many LLM unlearning methods. Performing GA is equivalent to doing Gradient Descent (GD) on the negative cross-entropy loss function, which is denoted as $\Loss_{\GA}(\Par)$ defined in Eq.~(\ref{eqn.ga.loss}). Based on gradient ascent, a large class of unlearning methods performs gradient-based optimization on a linear combination of the GA loss $\Loss_{\GA}$ and several other loss functions that encourage unlearning~\citep{jang2022knowledge,yao2023large, chen2023unlearn, maini2024tofu, eldan2023s}. Such a loss function can be written as 
\begin{equation}\label{eqn.general.ga.loss}
    \Loss(\Par) = \clga \Loss_{\GA}(\Par)
    + \clfg \Loss_{\FG}(\Par)
    + \clrt \Loss_{\RT}(\Par)
    - \ckfg \Lossdiv_{\FG}(\Par)
    + \ckrt \Lossdiv_{\RT}(\Par),
\end{equation}
where $\clga, \clfg, \clrt, \ckfg, \ckrt$ are non-negative weights. Here, $\Loss_{\FG}(\Par) = - \E_{\dsetf}[ \log(\policy_\Par(\tilde \y | \x))]$ is the Forget loss where $(\x, \y)\sim \dsetf$ and $\tilde\y\neq \y$ is any response for prompt $x$ which show some extent of ignorance towards the question $\x.$ $\Loss_{\RT}(\Par) = - \E_{\dsetr}[\log(\policy_\Par(\y |  \x))]$ is the retain loss. $\Lossdiv_{\FG}(\Par) = \E_{\dsetf}[ \kldivergence ( \policy_\Par(\cdot | \x) || \policy_\refm(\cdot |  \x)) ]$ is the expected KL divergence on the forget set. $\Lossdiv_{\RT}(\Par) = \E_{\dsetr}[ \kldivergence ( \policy_\Par(\cdot | \x) || \policy_\refm(\cdot | \x)) ]$ is the expected KL divergence on the retain set. In our experiments, we use three GA-based methods reported in \cite{maini2024tofu}, referred to as GA, GA+RT, GA+KL, which fall in this class of loss function. The weights in Eq.~(\ref{eqn.general.ga.loss}) are shown in \Cref{table.loss.coeff}.

{
\renewcommand{\arrayrulewidth}{1pt}
\begin{table}[H]
    \centering
    \begin{tabular}{cccccc}
    \toprule
        Loss & $\clga$ & $\clfg$ & $\clrt$ & $\ckfg$ & $\ckrt$\\
        \midrule
        GA & 1 & 0 & 0 & 0 & 0\\
        \midrule
        GA+RT & 1 & 0 & 1 & 0 & 0\\
        \midrule
        GA+KL & 1 & 0 & 0 & 0 & 1\\
        \midrule
        IDK+RT & 0 & 1 & 1 & 0 & 0\\
        \midrule
    \end{tabular}
    \caption{The weights for different components in GA-based loss functions and IDK+RT loss.}
    \label{table.loss.coeff}
\end{table}
}

\paragraph{IDK-based Methods ('I don't know').} \cite{maini2024tofu} proposed IDK+RT, which is a supervised loss function comprising of the retain loss and IDK loss term. The IDK loss term $\Loss_{\FG}$ is the averaged cross-entropy loss for question-answer pairs with questions $\x$ from the forget set $\dsetf$ and answers $\y$ replaced by $\tilde \y = $ 'I don't know' or a similar sentence showing ignorance towards this question. IDK+RT does not involve GA loss, and in general, IDK+RT loss shows a higher stability than GA-based methods. 

\paragraph{DPO-based Methods.} We also tested the DPO method \citep{rafailov2024direct} and its variants by adding either the retain loss or the KL divergence on the retain set. In the DPO loss, we take 'I don't know' or its variants as positive responses and the answers in the forget set as negative responses. We use $\beta=0.1$ in all DPO-based experiments, which is commonly recognized as the optimal inverse temperature in most cases. 

\paragraph{KTO-based Methods.} We examine Kahneman-Tversky Optimization (KTO) \citep{ethayarajh2024kto}, an alignment method with only non-paired preference data. The objective function of KTO is (we use a slightly different version than the original one as in \cite{ethayarajh2024kto})
\begin{equation}\label{eqn.KTO.loss}
    \Loss_{\mathrm{KTO}} := \frac{2}{\beta} \E_{\dsetf} \left[- \log \sigma \left(z_{\refm} - \beta \log \frac{\policy_{\Par} (\y \mid \x)}{\policy_{\refm}(\y \mid \x)} \right) \right],
\end{equation}
where $\beta > 0$ is the inverse-temperature, $\sigma$ is the sigmoid function, and
\begin{equation}
    z_{\refm} := \E_{x \sim \dsetf} \left[\beta \cdot \kldivergence \left(\policy_{\Par}(\cdot \mid \x) || \policy_{\refm}(\cdot \mid \x)\right)\right].
\end{equation}
Following \citep{ethayarajh2024kto}, we estimate the KL term via averaged log probability ratio for questions in the forget set and answers in the "I don't know" set (as the unrelated outputs). We examine both KTO and KTO+RT in our experiments with $\beta=0.1.$

\subsection{Full Results}\label{appendix.full.result}
\begin{figure}[t]
    \centering
    \includegraphics[width = 1\linewidth]{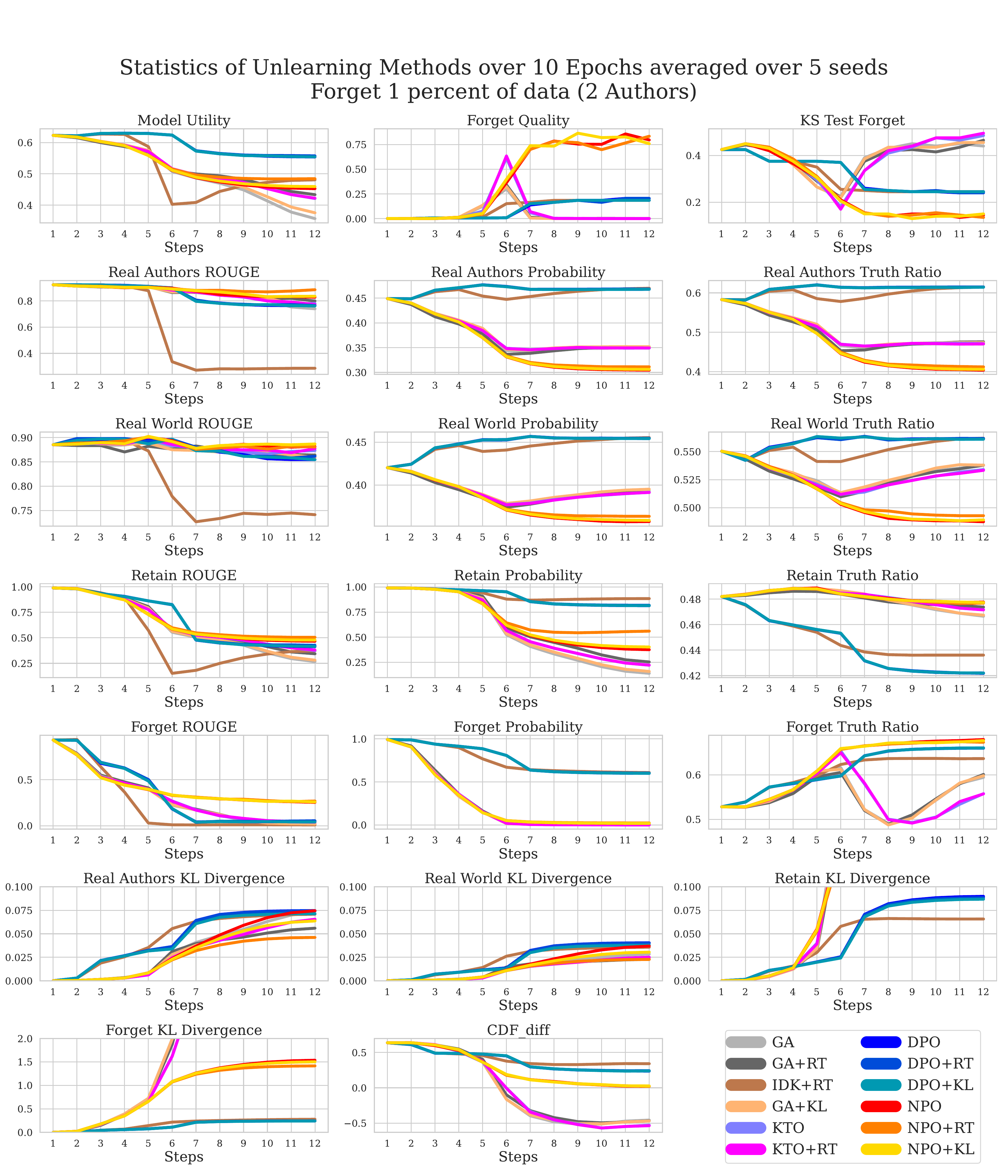}
    \caption{Statistics for NPO-based methods and baselines on the Forget01 task of TOFU.}
    \label{fig.forget01.full}
\end{figure}

\begin{figure}[t]
    \centering
    \includegraphics[width = 1\linewidth]{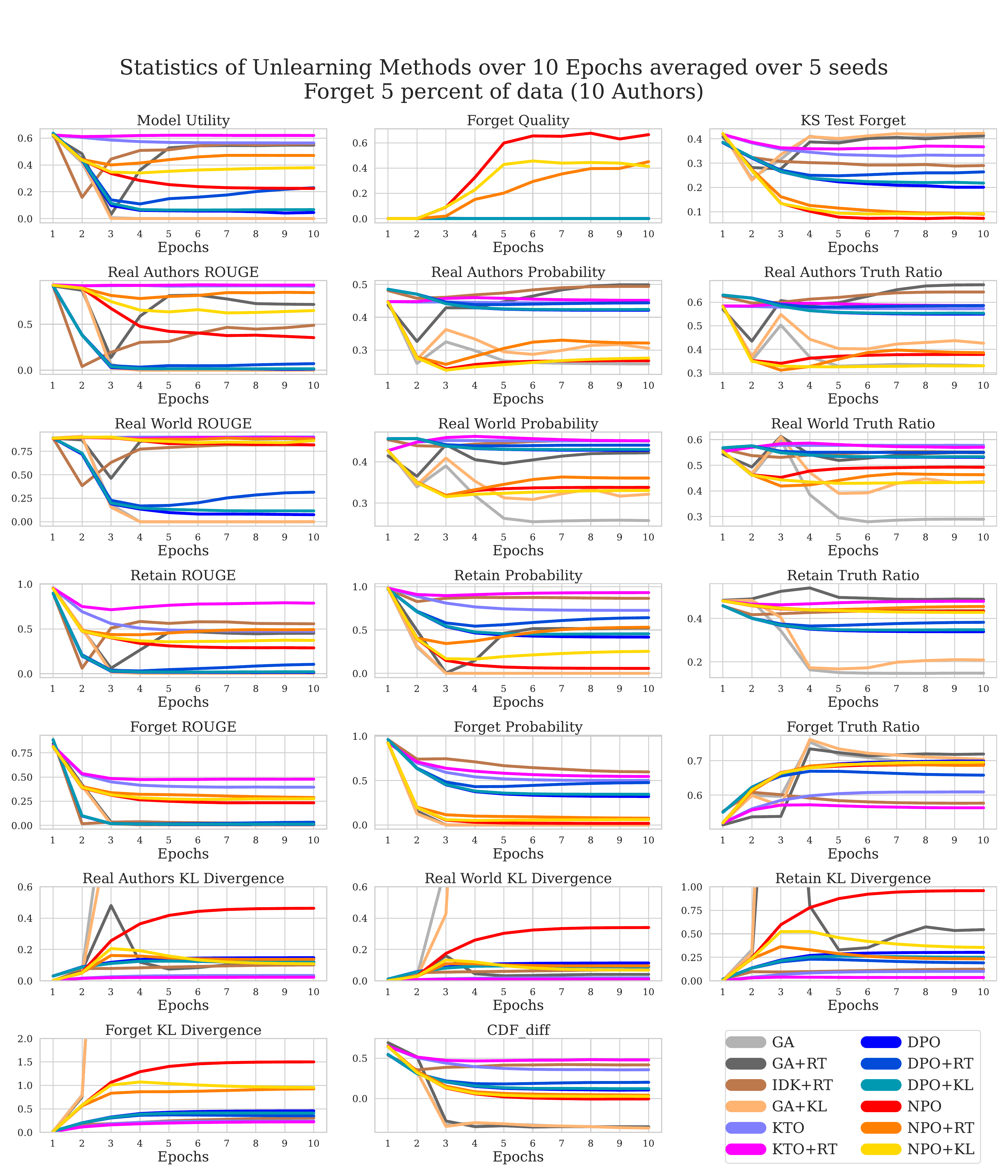}
    \caption{Statistics for NPO-based methods and baselines on the Forget05 task of TOFU.}
    \label{fig.forget05.full}
\end{figure}

\begin{figure}[t]
    \centering
    \includegraphics[width = 1\linewidth]{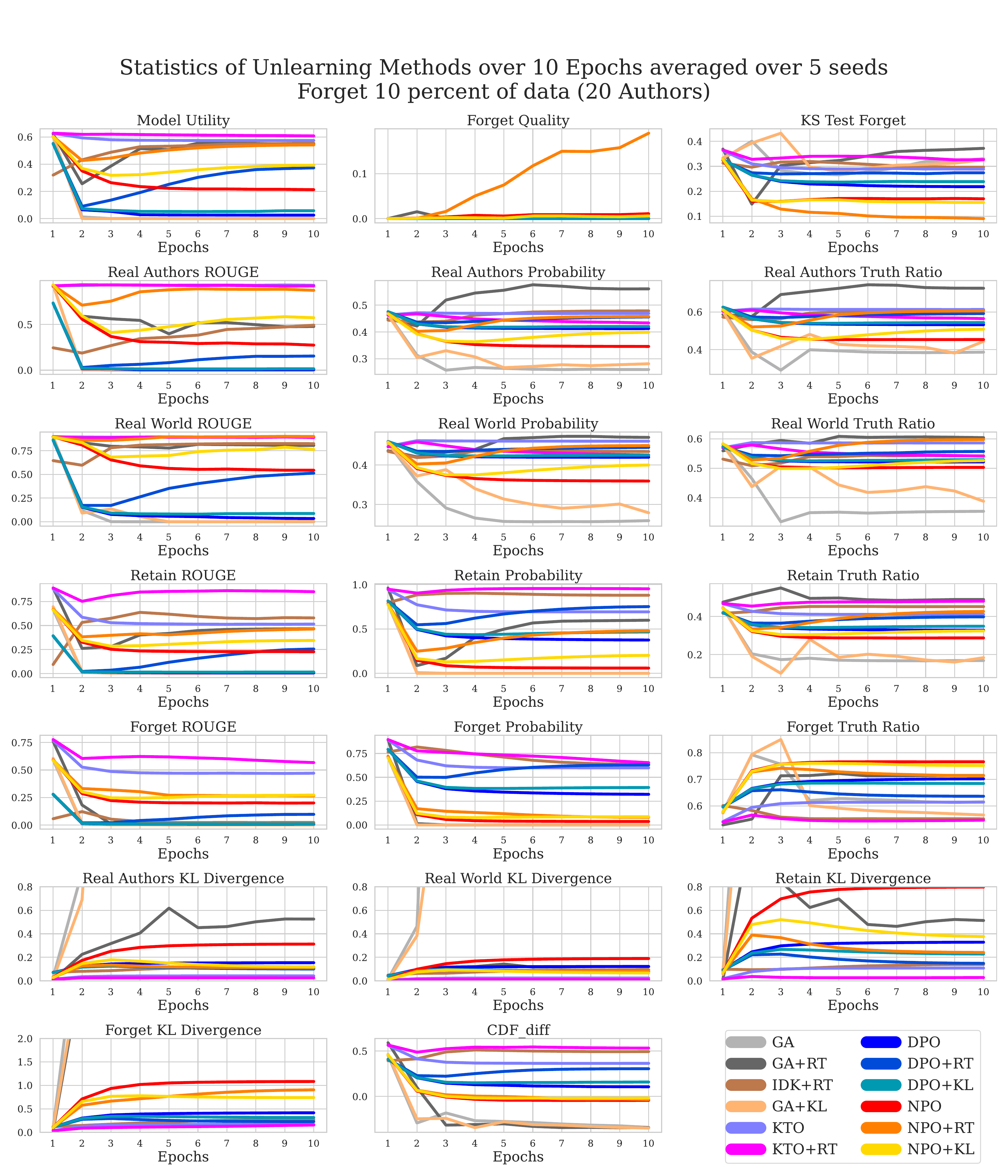}
    \caption{Statistics for NPO-based methods and baselines on the Forget10 task of TOFU.}
    \label{fig.forget10.full}
\end{figure}

\begin{figure}[t]
    \centering
    \includegraphics[width = 1\linewidth]{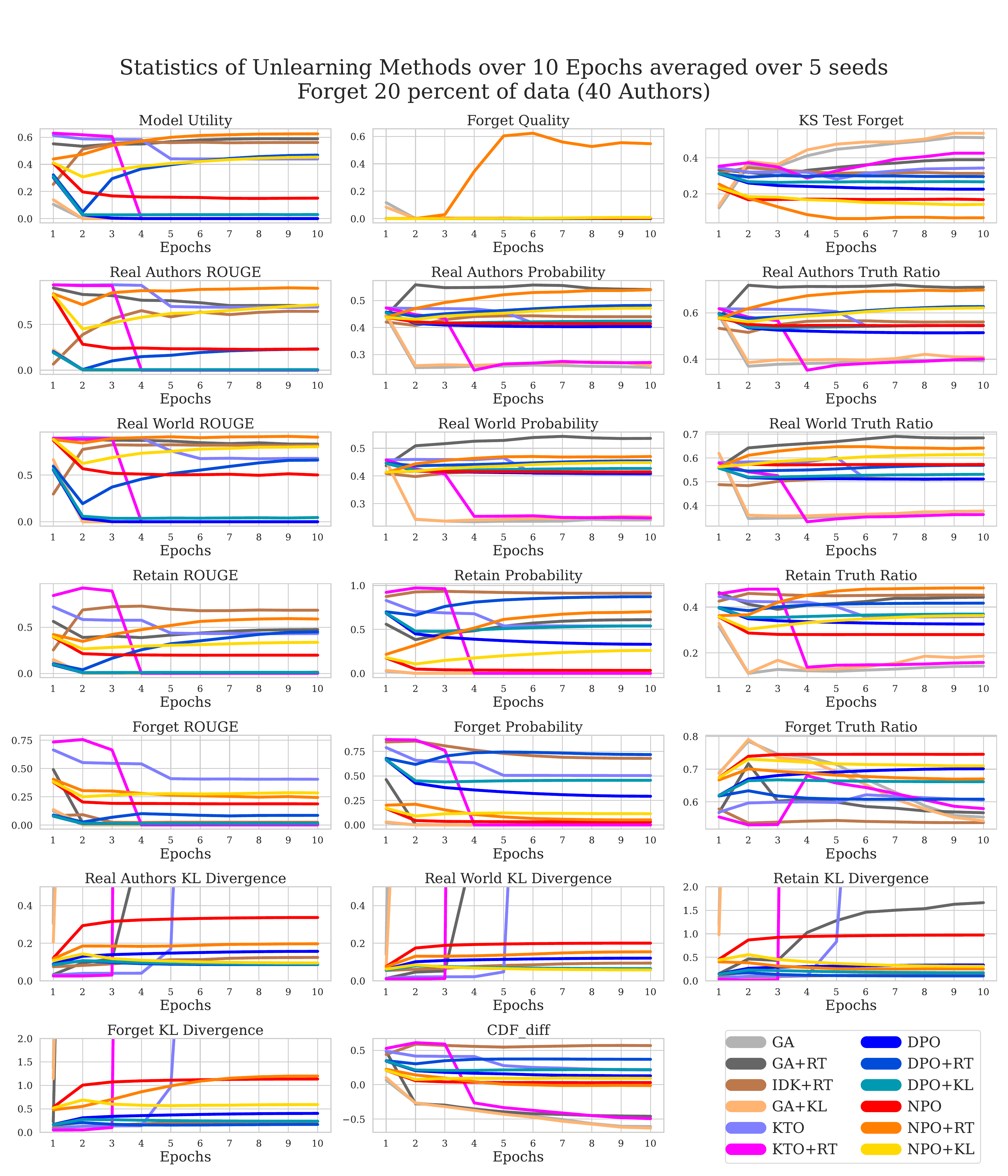}
    \caption{Statistics for NPO-based methods and baselines on the Forget20 task of TOFU.}
    \label{fig.forget20.full}
\end{figure}

\begin{figure}[t]
    \centering
    \includegraphics[width = 1\linewidth]{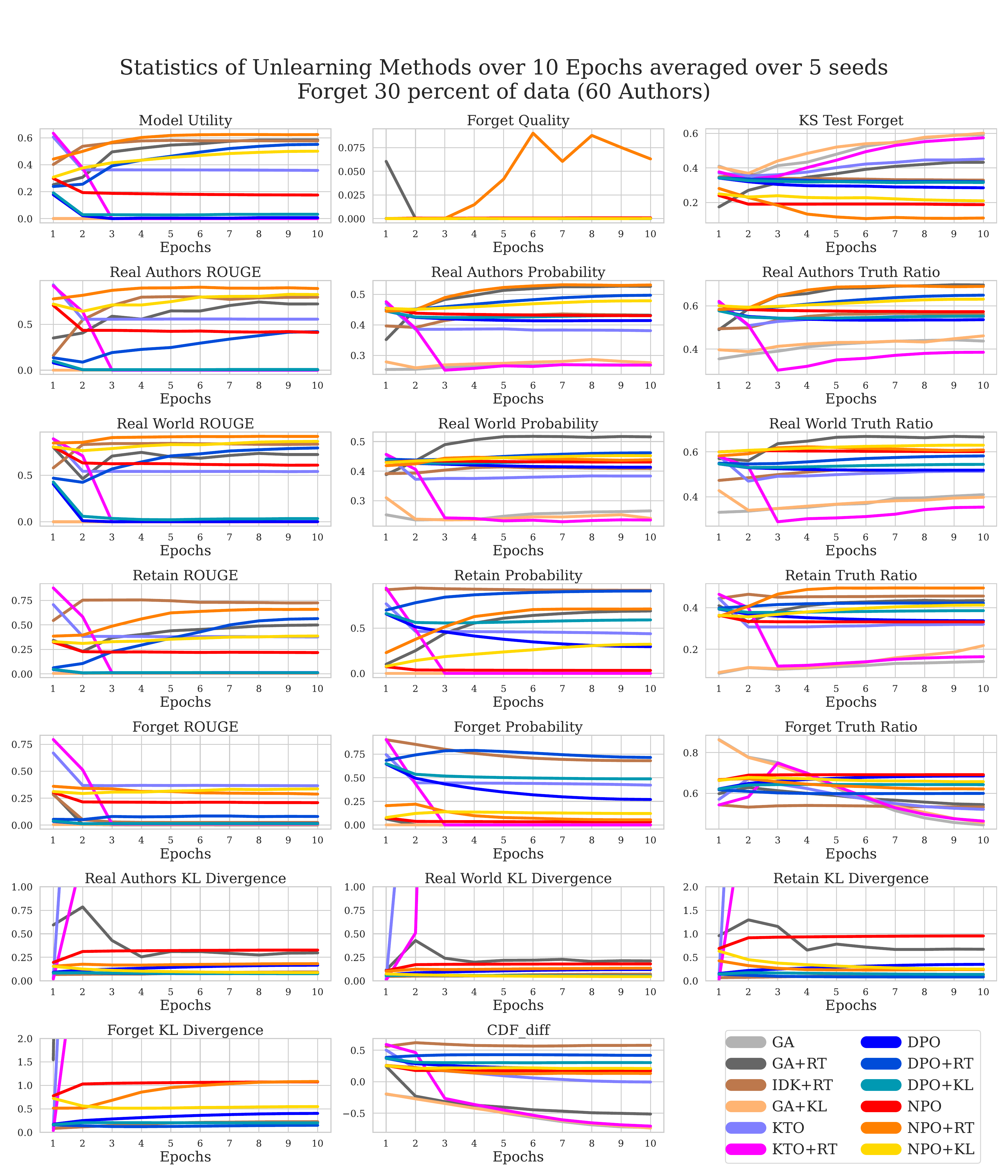}
    \caption{Statistics for NPO-based methods and baselines on the Forget30 task of TOFU.}
    \label{fig.forget30.full}
\end{figure}

\begin{figure}[t]
    \centering
    \includegraphics[width = 1\linewidth]{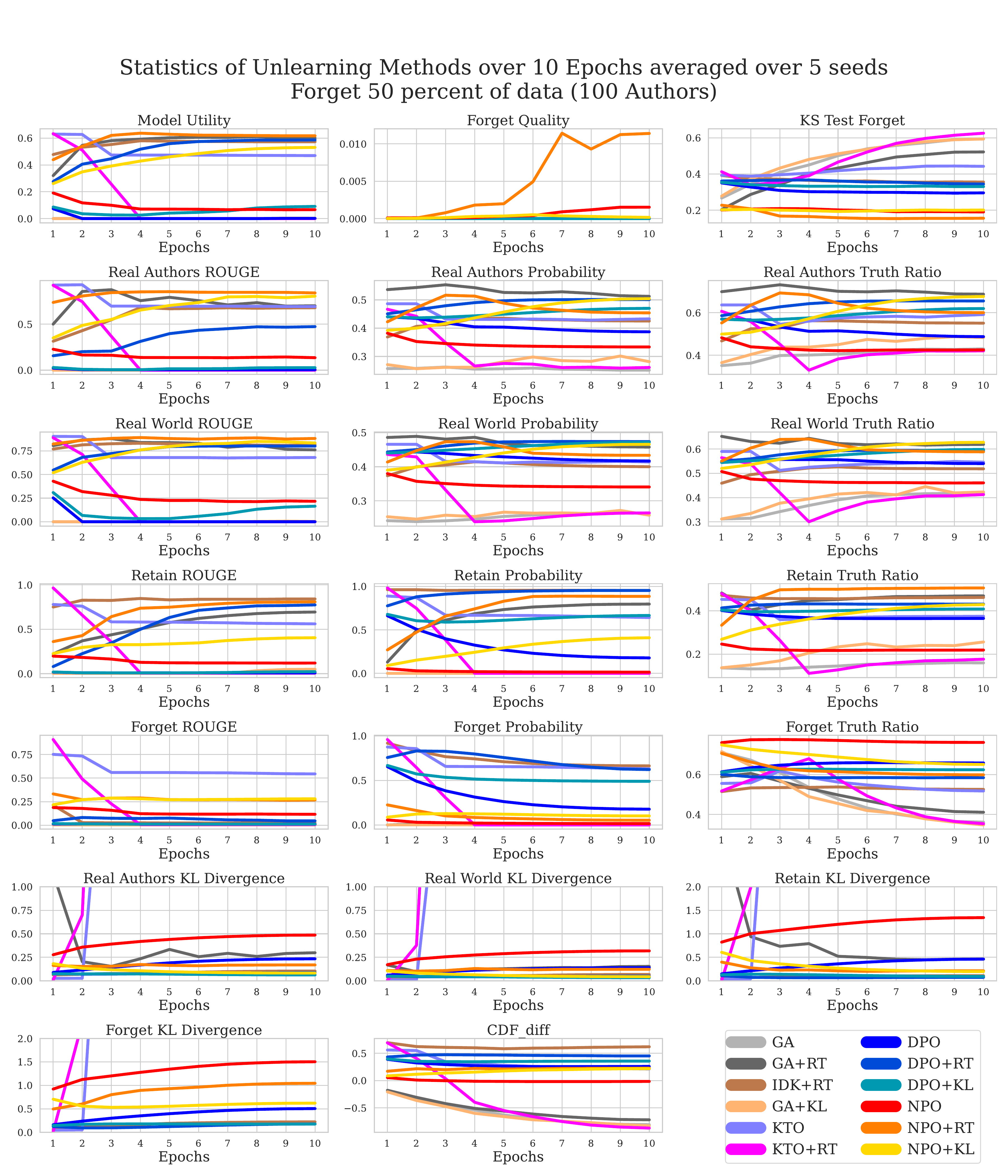}
    \caption{Statistics for NPO-based methods and baselines on the Forget50 task of TOFU.}
    \label{fig.forget50.full}
\end{figure}

\end{document}